\documentclass[10pt,twocolumn,letterpaper]{article}

\usepackage{iccv}
\usepackage{multirow}
\usepackage{diagbox}
\usepackage[table]{xcolor}
\usepackage{booktabs}
\usepackage{makecell}
\usepackage{placeins}
\usepackage[normalem]{ulem}

\usepackage{ulem}
\usepackage{soul}
\usepackage{enumitem} 
\definecolor{iccvblue}{rgb}{0.21,0.49,0.74}
\usepackage[pagebackref,breaklinks,colorlinks,allcolors=iccvblue]{hyperref}

\usepackage{amsmath}
\usepackage{amsthm}
\usepackage{graphicx}
\usepackage{tcolorbox}
\usepackage{multirow}
\usepackage{colortbl}
\usepackage{bbding}
\usepackage{wrapfig}
\usepackage{array}

\usepackage{amssymb,booktabs}
\usepackage{tabularx}
\usepackage{bbding}
\usepackage{pifont}
\usepackage{makecell}
\usepackage{float}
\usepackage{caption}

\def\paperID{*****}
\def\confName{ICCV}
\def\confYear{2025}

\title{ArtLang: Structured Language-to-Kinematics Grounding \\ for Articulated 3D Actuation}

\author{
\textbf{Sylvia Yuan}\textsuperscript{1,*},
\textbf{Dan Wang}\textsuperscript{1,*},
\textbf{Ravi Ramamoorthi}\textsuperscript{1},
\textbf{Xinrui Cui}\textsuperscript{2,\dag}
\\\\
\textsuperscript{1}University of California San Diego \quad
\textsuperscript{2}University of North Texas \\
\\
{\tt\small \textsuperscript{*}Equal Contribution \quad \textsuperscript{\dag}Corresponding Author
}
\\
{\tt\small Email: xinrui.cui@unt.edu}
}

\newcommand{\artlang}{ArtLang}

\begin{document}

\maketitle

\begin{abstract}
Articulated-object reconstructions recover explicit geometry and kinematics, but their parts often remain semantically anonymous and must be controlled through part indices and numerical joint parameters. We present \artlang{}, a framework for open-vocabulary language control of persistent reconstructed articulated assets. \artlang{} represents an asset as a semantic-kinematic articulation graph and augments its surface with language features and graph-constrained motion. Open-vocabulary proposals are bound to reconstructed parts while allowing uncertain parts to remain unnamed. A typed parser converts a command into a directive graph containing referring expressions, actions, magnitudes, reference frames, and relations. We then solve a global graph-to-graph grounding problem that jointly reasons about semantic, spatial, relational, and kinematic compatibility, with support for null assignments and abstention under ambiguity. Accepted directives are converted into continuous joint targets within the observed motion range and executed through forward kinematics. Experiments on synthetic reconstructions, mesh-based assets, and real captures demonstrate reliable language grounding and continuous articulated control across repeated parts, spatial references, relational commands, and ambiguous instructions.

\end{abstract}
\section{Introduction}
\label{sec}

\begin{figure}
\centering
\includegraphics[width=\linewidth]{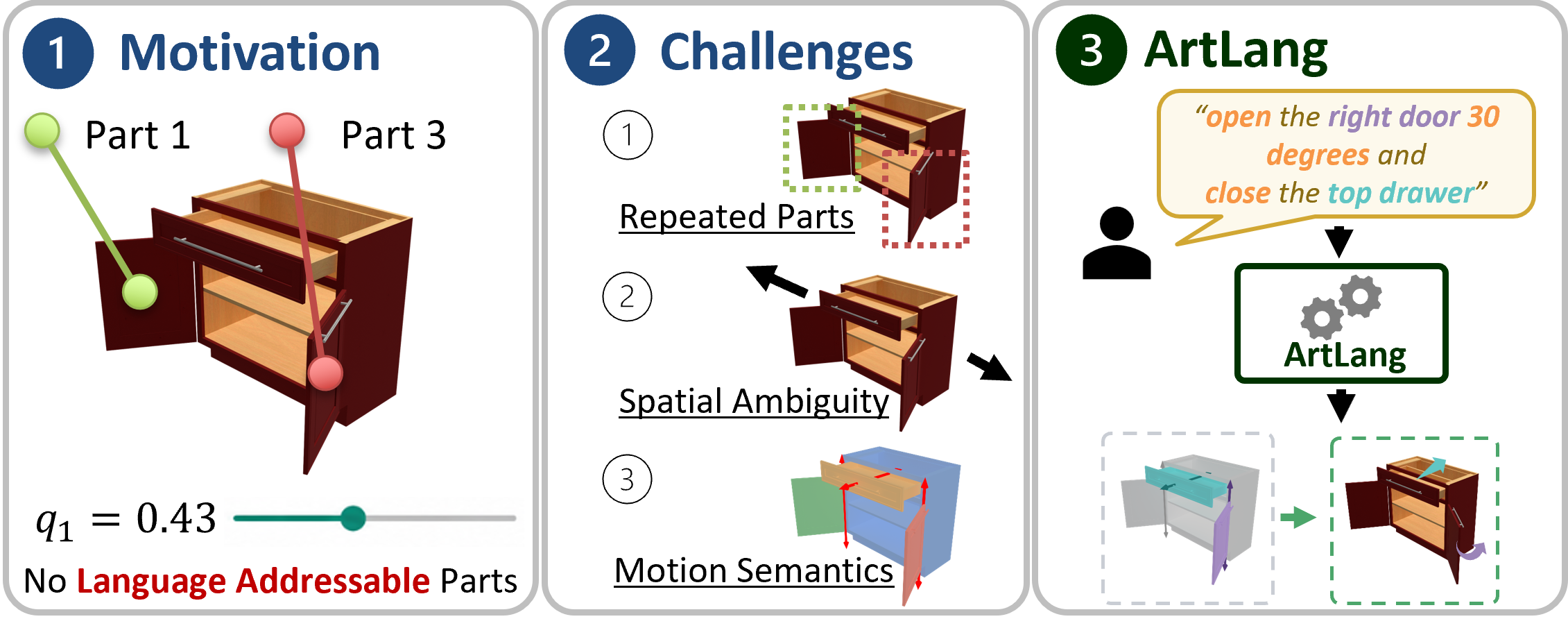}
\caption{\textbf{Motivation.} Reconstructed articulated assets lack a language-addressable interface. \artlang{} resolves semantic, spatial, and motion ambiguities to map commands to persistent parts and executable joint coordinates.}
\label{fig:motivation}
\end{figure}

Many objects in robotic manipulation, simulation, and digital twins are kinematic systems rather than rigid shapes: cabinets contain repeated doors, laptops rotate about hinges, and tools couple parts through joints. Recent methods reconstruct such objects as part-aware assets with explicit geometry and kinematics \cite{artmesh,artgs,gaussianart,articulate-anything}. Yet their parts remain \emph{semantically anonymous}: users must address them by index and control numerical joint parameters. A reconstruction may know how part~$3$ moves, but not that it is the \emph{right door} or should be \emph{all the way open} (Figure~\ref{fig:motivation}). This gap limits open-vocabulary inspection, editing, and control of reconstructed assets.

Open-vocabulary 3D language fields attach language features to spatial primitives, enabling text retrieval in static scenes \cite{lerf,langsplat}. Semantic localization alone does not provide articulation control: localizing \emph{door} does not identify the intended instance, resolve whether \emph{right} is camera-, user-, or object-relative, or map \emph{open} to a valid joint coordinate. Existing language-driven articulation methods mainly generate or infer articulated models rather than bind commands to reconstructed parts and continuous coordinates \cite{lam, atop}. The missing capability is structured binding from language to an asset's semantic and kinematic structure, mapping commands such as \emph{open the right door} to a posed 3D object (Figure~\ref{fig:motivation}).

This binding is difficult because semantic, spatial, and kinematic ambiguities are coupled. Repeated parts may share appearance and semantics; spatial descriptions are frame-dependent and may include reference-only entities, as in \emph{open the left door below the drawer}; and multiple directives must be resolved jointly to satisfy relations and avoid conflicts. Motion expressions such as \emph{halfway open}, \emph{open by $20^\circ$}, and \emph{open to $20^\circ$} also differ and depend on joint type, current state, and observed motion. Because reconstruction, vision-language proposals, and commands are uncertain, a reliable system should expose ambiguity rather than silently choose a part or alter the action.

We present \textbf{\artlang{}}, which (i) identifies the referred persistent part, (ii) selects its executable joint, (iii) infers the requested joint coordinate, (iv) executes the resulting kinematic state, given a reconstructed articulated asset and a compositional natural-language command. ArtLang jointly formulates language-conditioned actuation and articulation control as \textbf{structured grounding graph-to-graph reasoning and inference}.
The reconstructed asset is an articulation graph $\mathcal A=(\mathcal V,\mathcal E)$ whose nodes are rigid parts and whose edges store recovered one-degree-of-freedom motion. A typed parser converts the command into a directive graph $\mathcal D=(\mathcal V_D,\mathcal E_D)$ containing referring expressions, actions, magnitudes, temporal stages, reference frames, and pairwise relations; nodes may be actionable or reference-only.

Rather than resolving phrases independently, \artlang{} jointly assigns directive nodes to part or private null nodes using semantic, spatial, relational, and action-joint compatibility. This preserves shared landmarks, prevents conflicting actions, and permits abstention when the selected interpretation is not better than competing global assignments.

The graph formulation is coupled to a\textbf{ factorized semantic-kinematic field} on the canonical surface. The factorized field $\Phi(x)=(\pi(x),f_\theta(x),\mathcal P(\pi(x)))$, where $\pi(x)$ encodes part identity, $f_\theta(x)$ provides language features, and $\mathcal P(\pi(x))$ represents graph-constrained motion, records each point's persistent part, dense open-vocabulary feature, and kinematic path.
During \textbf{open-vocabulary part binding}, we distill multi-view language features onto the surface and bind vision-language proposals to reconstructed parts. Private dummy matches allow unnamed parts, avoiding the assumption of one correct proposal per part while preserving repeated noun instances.

At the \textbf{directive graph and structured grounding} during query time, each referring-expression embedding is compared with $f_\theta(x)$ across the surface; the resulting evidence is aggregated per part and combined with explicit-frame geometry, pairwise relations, and action-joint compatibility in global grounding.

For each accepted action, a typed operator converts the verb and magnitude into a selected-joint coordinate. It distinguishes endpoint requests, absolute and relative fractions, and angular or translational changes.
Accepted targets are composed by forward kinematics, yielding a consistent configuration or motion clip rather than an appearance-only edit.

The system binds dense language features to persistent parts, resolves relations globally, and maps accepted actions to executable joint coordinates with explicit failure. We evaluate on synthetic multi-view reconstruction with Articulate-100, a given-mesh setting isolating language grounding from geometry recovery, and real SplArt captures.

Our contributions are:
\begin{itemize}
    \item We formulate language-conditioned control of a reconstructed articulated asset as structured grounding from a directive graph to a  persistent semantic-kinematic articulation graph, with explicit part,  relation, joint, and uncertainty variables.
    \item We introduce robust semantic-geometric binding that combines a dense geometry language field and per-view semantic-geometric partial matching, without assuming one correct proposal per part.
    \item We develop a globally consistent grounding and actuation method that combines semantic, explicit-frame spatial, relational, and action-joint compatibility, then maps accepted directives to typed continuous targets inside the observation-supported joint interval with calibrated abstention.
    \item An evaluation protocol with metrics that score the semantic, spatial, and motion axes independently, together with experiments across synthetic multi-view reconstruction on Articulate-100, a mesh-input setting where the geometry is given, and real captures on SplArt.
\end{itemize}

\section{Related Work}
\label{sec:related}

\emph{\textbf{Reconstructing articulated objects.}}
NeRF- and SDF-based methods \cite{nerf,neus,asdf,cla-nerf,paris,digitaltwinart,articulate-nerf} recover implicit fields that require surface extraction. Gaussian methods \cite{3dgs,artgs,gaussianart,splart,part2gs,articulated-gs,reartgs} attach rigid or screw motion to point primitives without connectivity. Mesh-based differentiable rendering and splat-to-surface methods \cite{2dgs,rade-gs,sugar,milo,triangle-splatting,meshsplatting} culminate in ArtMesh \cite{artmesh}, which reconstructs connected part-aware meshes with rigid trajectories. We use ArtMesh as the reconstruction backbone and focus on language-based part selection and actuation.

\emph{\textbf{Estimating and generating articulation.}}
Feedforward methods estimate part masks, motion parameters, or kinematic structure from images and priors \cite{shape2motion,captra,sage,real2code,urdformer,larm,opd,opdmulti,partrm}. Generative methods synthesize articulated assets \cite{cage,singapo,meshart,artilatent,geopard,freeart3d,dreamart} or articulate existing meshes \cite{articulate-anymesh}. Language and vision-language models generate URDFs \cite{articulate-anything,kinematify,urdfanything}, predict affordances \cite{a3vlm,manipllm}, create articulated geometry from text \cite{lam}, and personalize motion \cite{atop}. These methods construct or infer articulation rather than provide a runtime language interface to a persistent reconstruction.

\emph{\textbf{Language grounding in 3D scenes.}}
Language features have been distilled into radiance fields \cite{dff,lerf}, Gaussian primitives \cite{langsplat,feature3dgs}, scene representations \cite{openscene,clipfields}, and hierarchical groups \cite{garfield}; instruction-driven methods edit appearance \cite{in2n,gaussianeditor}. These systems localize concepts or modify surfaces in static scenes but do not disambiguate repeated articulated parts or map verbs to joint motion. \artlang{} adds explicit spatial disambiguation and executable articulation trajectories.

\emph{\textbf{Language-driven control of articulated objects.}}
Prior language-based methods typically operate during asset creation, after which parts are controlled by joint indices. Affordance systems \cite{where2act,a3vlm,manipllm} instead ground instructions to interaction points for manipulation policies, not recovered joints. In contrast, \artlang{} uses free-form runtime language to select a persistent part, resolve its spatial instance, and actuate it along its recovered joint.

\section{Method}
\label{sec:method}

\begin{figure*}[t]
  \centering
  \includegraphics[width=\textwidth]{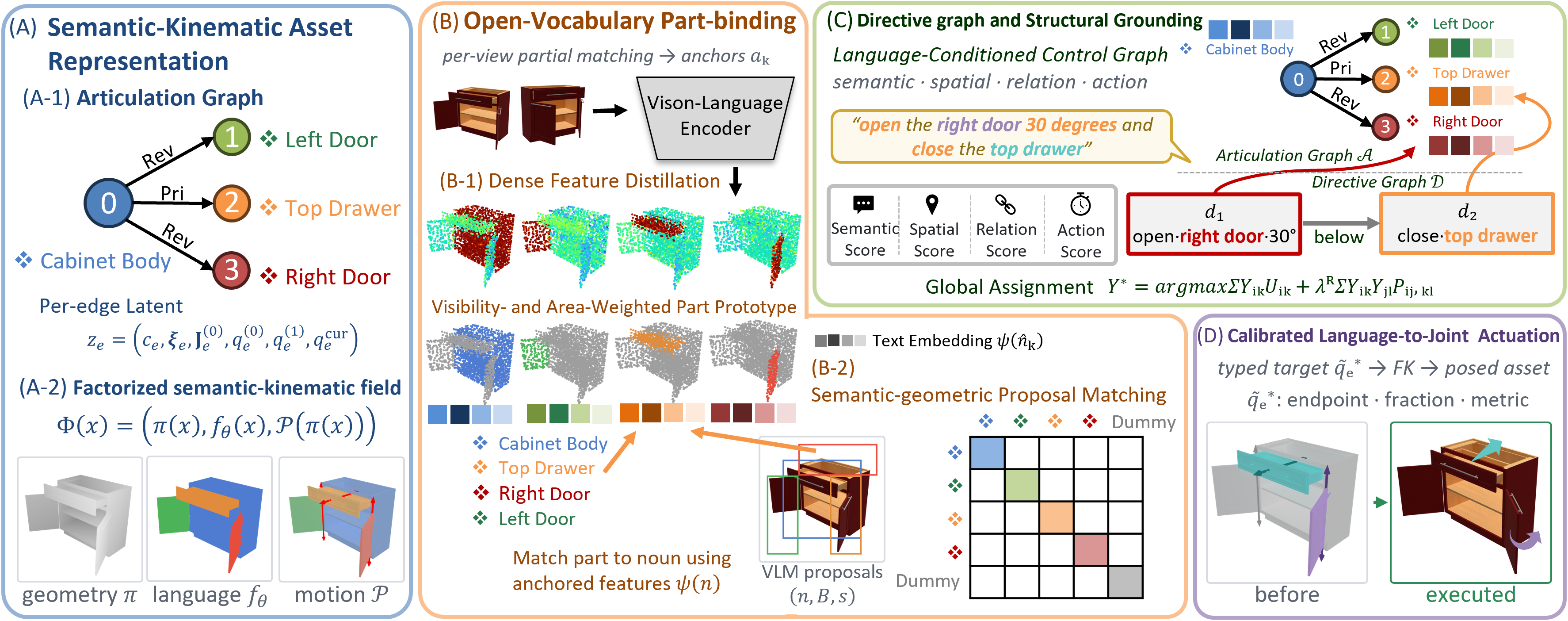}
 \caption{\textbf{Overview of \artlang{}.} \textbf{(A)} A part-aware reconstruction is represented by an articulation graph and factorized semantic-kinematic field. \textbf{(B)} Open-vocabulary features and proposals are bound to persistent parts by semantic-geometric partial matching. \textbf{(C)} A directive graph is globally grounded using semantic, spatial, relational, and action compatibility. \textbf{(D)} Accepted directives are converted to joint targets and executed by forward kinematics. The resulting graph-to-graph pipeline converts compositional language into globally consistent and executable articulation. }
  \label{fig:arch}
\end{figure*}

\textbf{Overview.}
\artlang{} grounds a compositional command in the persistent structure of a reconstructed articulation asset.
\artlang{} couples three components: an \emph{asset graph} of reconstructed parts and their recovered joints, a \emph{language field} that attaches open-vocabulary features to the asset's surface, and a \emph{directive graph} holding the entities, actions, magnitudes, frames, and relations of a command.
We propose a \textit{graph-to-graph binding} from language to an articulated object's semantic and kinematic structure. We bind visual-language proposals to part nodes by partial matching, then solve a global directive-to-part assignment under semantic, spatial, relational, and action constraints. Accepted action nodes are converted into continuous joint coordinates within the observation-supported motion interval. This formulation addresses repeated parts, frame-dependent language, multi-directive consistency, kinematic calibration, and ambiguous commands in one structured model.

\subsection{Semantic-Kinematic Asset Representation}
\label{sec:asset_representation}
\textbf{Articulation graph.}
This stage corresponds to panel~(A) of Figure~\ref{fig:arch}. The input is either multi-view observations at two articulation states or a rest-state mesh with observations of a second state. Previously, articulated reconstruction works such as ArtGS, GaussianArt, and ArtMesh\cite{artgs, gaussianart, artmesh} return motion parameters and reconstructed representation. Instead, we construct an articulation graph. We adopt a part-aware reconstruction backend ArtMesh~\cite{artmesh}, returns
\begin{equation}
\begin{aligned}
  \mathcal A&=\bigl(\mathcal V,\mathcal E,
  \{\bar{\mathcal G}_k\}_{k=0}^{K-1},\{z_e\}_{e\in\mathcal E}\bigr),\\
  z_e&=(c_e,\boldsymbol\xi_e,\mathbf J_e^{(0)},
  q_e^{(0)},q_e^{(1)},q_e^{\rm cur}).
\end{aligned}
  \label{eq:asset_graph}
\end{equation}
Here, node $k$ is a rigid part with local geometry $\bar{\mathcal G}_k$, and edge $e=(p(k),k)$ is a revolute, prismatic, or fixed joint. The unit twist $\boldsymbol\xi_e\in\mathbb R^6$ is expressed in the parent frame, $\mathbf J_e^{(0)}$ is the parent-to-child transform at the first observed state, and $q_e^{(0)},q_e^{(1)}$ are the recovered normalized ($q_e^{(0)}{=}0$ at the first observed state, $q_e^{(1)}{=}1$ at the second) coordinates of the two observations, so that $\mathcal I_e=[0,1]$; the physical joint magnitude, the full swing angle or translation of each part, is recovered alongside and used to map metric requests such as an explicit angle onto this normalized interval. They define the supported interval
\begin{equation}
  \mathcal I_e=
  [\min(q_e^{(0)},q_e^{(1)}),\max(q_e^{(0)},q_e^{(1)})].
  \label{eq:supported_interval}
\end{equation}
We do not interpret $\mathcal I_e$ as the object's complete mechanical range. When the reconstruction backend does not estimate a hierarchy, $\mathcal A$ is a star graph rooted at the static base. Using the matrix representation $\widehat{\boldsymbol\xi}_e\in\mathfrak{se}(3)$, the parent-to-child transform at coordinate $q_e$ is
\begin{equation}
  \mathbf J_e(q_e)=
  \exp\!\left((q_e-q_e^{(0)})\widehat{\boldsymbol\xi}_e\right)
  \mathbf J_e^{(0)}.
  \label{eq:joint_transform}
\end{equation}
For a revolute joint with unit axis $\mathbf u$ through point $\mathbf p$, we use the spatial-twist convention $\boldsymbol\xi=(-\mathbf u\!\times\!\mathbf p,\mathbf u)$; for a prismatic joint, $\boldsymbol\xi=(\mathbf u,\mathbf 0)$. The world transform of part $k$ and the posed asset are
\begin{equation}
\begin{aligned}
  \mathbf T_0(\mathbf q)&=\mathbf I,\\
  \mathbf T_k(\mathbf q)&=
  \mathbf T_{p(k)}(\mathbf q)
  \mathbf J_{(p(k),k)}(q_{(p(k),k)}),\\
  \mathcal G(\mathbf q)&=
  \bigcup_k\mathbf T_k(\mathbf q)\bar{\mathcal G}_k.
\end{aligned}
  \label{eq:forward_kinematics}
\end{equation}
The per-part motion is recovered from the two observed states: with multi-view input we adopt the ArtMesh articulation optimization, while with mesh input, where both states are provided as meshes, each joint is recovered by a correspondence-free registration of the two states.

\textbf{Factorized semantic-kinematic field.}
Let $\mathbf q^0$ denote the canonical configuration and $\mathcal G_k=\mathbf T_k(\mathbf q^0)\bar{\mathcal G}_k$. On the canonical surface $\Omega=\bigcup_k\mathcal G_k$, let $\pi:\Omega\rightarrow\mathcal V$ assign each point to its reconstructed part. We learn a dense language feature $f:\Omega\rightarrow\mathbb R^{d}$ and bind each point to the joints on the kinematic path of its part. For a canonical point $x\in\mathcal G_k$, define its kinematic map
\begin{equation}
  \mathcal P(\pi(x))=
  \mathbf T_k(\mathbf q)
  \mathbf T_k(\mathbf q^{0})^{-1}\tilde x,
  \qquad k=\pi(x),
  \label{eq:point_kinematic_map}
\end{equation}
where $\tilde x$ is the homogeneous coordinate of $x$. The representation is the articulation-structured language field
\begin{equation}
  \Phi(x)=\bigl(\pi(x),f_\theta(x),\mathcal P(\pi(x))\bigr),
  \qquad x\in\Omega,
  \label{eq:factorized_field}
\end{equation}
Geometry is the domain $\Omega$; the three returned attributes encode part identity, language features, and graph-constrained motion. Eq.~\eqref{eq:factorized_field} attaches the semantic and executable kinematic attributes needed for language control.

\subsection{Open-Vocabulary Part Binding}
\label{sec:part_binding}
A reconstructed graph provides stable part instances but no natural-language labels. A vision--language model provides image semantics but does not automatically attach each proposal to a persistent 3D part. This stage transfers image-language understanding to the reconstructed surface, then binds proposal instances to graph nodes.
Every reconstructed part should carry a noun. A VLM will propose nouns, but it does not propose exactly one correct noun per part. It invents unreconstructed parts, misses others, and repeats itself across identical siblings. Binding must therefore be partial. Panel~(B) of Figure~\ref{fig:arch} illustrates this stage.

\textbf{Dense feature distillation.}
For view $c$, a frozen vision-language encoder produces a dense feature map $\mathbf F_c(\mathbf u)\in\mathbb R^{d}$. Rendering the field gives
\begin{equation}
\begin{aligned}
  \widehat{\mathbf F}_c(\mathbf u)
  &=\operatorname{norm}\!\left(
  \sum_{x\in\mathcal R(c,\mathbf u)}
  w_{c\mathbf u x}f_\theta(x)\right),\\
  \mathcal L_{\rm field}
  &=\frac{1}{|\mathcal O|}
  \sum_{(c,\mathbf u)\in\mathcal O}
  \bigl[1-\langle\widehat{\mathbf F}_c(\mathbf u),
  \mathbf F_c(\mathbf u)\rangle\bigr],
\end{aligned}
  \label{eq:field_distillation}
\end{equation}
where $w_{c\mathbf u x}$ are differentiable rendering weights. The loss is a cosine distance over a sampled set $\mathcal{O}$ of (view, pixel) pairs.

Dense features are useful for point-level localization but can be noisy on occluded or weakly observed surface regions. We therefore aggregate the dense feature into a visibility- and area-weighted part prototype
\begin{equation}
  \bar{\mathbf f}_k=
  \operatorname{norm}\!\left(
  \frac{\int_{\mathcal G_k}w(x)f_\theta(x)\,\mathrm dA}
       {\int_{\mathcal G_k}w(x)\,\mathrm dA+\epsilon}
  \right).
  \label{eq:part_prototype}
\end{equation}
The dense field supports point-level localization, while $\bar{\mathbf f}_k$ provides a stable node-level descriptor.

\textbf{Semantic-geometric proposal matching.}
A vision-language model proposes for the object a set of open-vocabulary part candidates $\{p_j=(n_j,B_j)\}_{j=1}^{M}$, each a part phrase $n_j$ and an image box $B_j$, in a single reference view. We project every reconstructed part $k$ into that view and take its image-space centroid $\mathbf q_k$; let $\mathbf b_j$ be the center of box $B_j$. Because the model already semantically localizes the candidates, we bind parts to candidates by projected geometry,
\begin{equation}
 C_{kj}=\frac{\|\mathbf q_k-\mathbf b_j\|_2^2}{d^2},
 \label{eq:proposal_cost}
\end{equation}
where $d$ is the image diagonal. The static base is pinned to the model's base candidate, and the remaining parts are assigned by a Hungarian solve; appending a private dummy candidate to each part relaxes the assignment so a part may remain unnamed when no candidate is compatible,
\begin{equation}
  \mathbf X^*=\arg\min_{\mathbf X\in\mathcal U}
   \sum_{k\in\mathcal V}\sum_{j=1}^{M+|\mathcal V|}X_{kj}\,C_{kj}.
  \label{eq:partial_matching}
\end{equation}
A matched part adopts its candidate phrase $\hat n_k=n_j$ and takes the semantic anchor
\begin{equation}
 \mathbf a_k=\psi(\hat n_k),
 \label{eq:part_anchor}
\end{equation}
the FG-CLIP2~\cite{fgclip2} text embedding of the assigned noun, with spatial qualifiers folded in at query time; an unmatched part keeps a generic label rather than a wrong noun. The binding is semantic through the open-vocabulary candidates it commits to parts and geometric through the projected match.

\subsection{Directive Graph and Structured Grounding}
\label{sec:structured_grounding}
This stage converts a natural-language command into a \emph{globally consistent} correspondence between linguistic entities and persistent parts of the reconstructed articulation graph. It infers (i) a possibly null directive-to-part assignment and (ii) a valid target coordinate for each accepted joint.

Panel~(C) of Figure~\ref{fig:arch} depicts the directive graph and its global assignment to the articulation graph.

\textbf{Language-conditioned control graph.}
A typed parser converts command $u$ into $\mathcal D=(\mathcal V_D,\mathcal E_D)$. Each entity node is
\begin{equation}
  d_i=(r_i,v_i,m_i,F_i,s_i,\eta_i),
  \label{eq:directive_node}
\end{equation}
where $r_i$ is the referring expression, $v_i$ the action, $m_i$ its typed magnitude, $F_i$ an explicit camera-, user-, or object-centric reference frame, $s_i\in\{1,2,\ldots\}$ the temporal executed stage, and $\eta_i\in\{0,1\}$ indicates whether the entity is actionable. Reference-only nodes $\eta_i=0$ support commands like \emph{open the drawer below the left door}: the door constrains the drawer but is not actuated. Each edge $(i,j,r)\in\mathcal E_D$ specifies a relation $r$, such as \emph{left of}, \emph{above}, \emph{below}, \emph{nearest}, or \emph{attached to}.

\textbf{Semantic compatibility.}
Let normalized text embedding $\mathbf t_i=\psi(r_i)$ and let $A_k$ be the surface area of part $k$. Semantic compatibility combines the stable part anchor with query-dependent dense evidence:
\begin{equation}
\begin{aligned}
  s^{\rm sem}_{ik}={}&
  \beta\langle\mathbf t_i,\mathbf a_k\rangle\\
  &+(1-\beta)\tau_f\log\!\left[
  \frac{1}{A_k}\int_{\mathcal G_k}
  \exp\!\left(
  \frac{\langle\mathbf t_i,f_\theta(x)\rangle}{\tau_f}
  \right)\mathrm dA\right].
  \label{eq:semantic_score}
\end{aligned}
\end{equation}
The first term provides stable part-level semantics, while the log-sum-exp term preserves localized dense-field evidence.

\textbf{Spatial compatibility.}
Suppose $r_i$ contains a directional modifier, e.g., \emph{left}, \emph{right}, \emph{above}, \emph{below}, \emph{front}, or \emph{back}. The parser maps the modifier to a signed unit vector $\mathbf d_i^{F_i}\in\mathbb R^3$ in the frame $F_i$ under a common right-handed convention. We transform the canonical part centroid of part $k$ into frame $F_i$ as $\boldsymbol\mu_k^F$ and define a soft semantic candidate distribution and its semantic center
\begin{equation}
\begin{aligned}
  \omega_{ik}&=\operatorname{softmax}_{k}
  (s^{\rm sem}_{ik}/\tau_s),\\
  \bar{\boldsymbol\mu}_i^{F_i}
  &=\sum_k\omega_{ik}\boldsymbol\mu_k^{F_i}.
\end{aligned}
  \label{eq:semantic_center}
\end{equation}
Spatial compatibility normalizes the signed displacement of part $k$ from center $\bar{\boldsymbol\mu}_i^{F_i}$ over all candidate parts,
\begin{equation}
\begin{aligned}
   s^{\rm sp}_{ik}
  &=\frac{(\mathbf d_i^{F_i})^\top
  (\boldsymbol\mu_k^{F_i}-\bar{\boldsymbol\mu}_i^{F_i})}
  {\max_\ell|(\mathbf d_i^{F_i})^\top
  (\boldsymbol\mu_\ell^{F_i}-\bar{\boldsymbol\mu}_i^{F_i})|+\epsilon}.
\end{aligned}
  \label{eq:spatial_score}
\end{equation}
For directives without a directional modifier, we set $s^{\rm sp}_{ik}=0$. Otherwise, Eq.~\eqref{eq:spatial_score} evaluates the modifier over a soft semantic candidate set rather than committing to a noun before spatial reasoning.

\textbf{Pairwise relational compatibility.} For relation $r$ between directives $i$ and $j$, we form the frame-conditioned geometric feature
\begin{equation}
\begin{aligned}
  \boldsymbol\rho_{k\ell}^{F}
  =\Bigl[&
  (\boldsymbol\mu_k^{F}-\boldsymbol\mu_\ell^{F})/d_{\rm obj},\\
  &\log\!\frac{\operatorname{size}(\mathbf B_k^{F})}
                  {\operatorname{size}(\mathbf B_\ell^{F})},
  \operatorname{IoU}(\mathbf B_k^{F},\mathbf B_\ell^{F}),
  d_{\mathcal A}(k,\ell)\Bigr],\\
  P_{ij,k\ell}&=g_r(\boldsymbol\rho_{k\ell}^{F_{ij}}),
\end{aligned}
  \label{eq:relation_score}
\end{equation}
where $g_r$ is a relation-specific scorer, $d_{\rm obj}$ normalizes object scale, and $d_{\mathcal A}$ is articulation-graph distance.

\textbf{Action compatibility.}
Let $\mathbf j_k$ contain the incident joint type, axis, and observed motion span. Action compatibility is
\begin{equation}
  s^{\rm act}_{ik}=
  \begin{cases}
    0, & \eta_i=0,\\
    g_{\rm act}\bigl(\psi(v_i),\mathbf j_k\bigr),
      & \eta_i=1,\ \chi(v_i,c_{e(k)})=1,\\
    -\infty, & \eta_i=1,\ \chi(v_i,c_{e(k)})=0,
  \end{cases}
  \label{eq:action_score}
\end{equation}
where the hard mask $\chi$ rejects fixed parts and unit--joint or verb--joint mismatches. For the root node, we set $c_{e(0)}=\mathrm{fixed}$. $U_{ik}$ is the unary term of the assignment objective in Eq.~\eqref{eq:structured_assignment}; the pairwise term adds relations. The complete unary score is
\begin{equation}
  U_{ik}=
  \lambda_{\rm sem}s^{\rm sem}_{ik}
  +\lambda_{\rm sp}s^{\rm sp}_{ik}
  +\lambda_{\rm act}s^{\rm act}_{ik}.
  \label{eq:unary_score}
\end{equation}
Each directive has a private null node. Its score depends on parser confidence and semantic-score entropy, providing an explicit outcome for unsupported or under-specified language.

\textbf{Global assignment.}
Let $Y_{ik}\in\{0,1\}$ assign directive $i$ to real part $k$ or to its private null node $\varnothing_i$. We solve
\begin{equation}
\begin{aligned}
  \mathbf Y^*=\arg\max_{\mathbf Y\in\mathcal Y}\quad
  &\sum_i\sum_{k\in\mathcal V\cup\{\varnothing_i\}}Y_{ik}U_{ik}\\
  &+\lambda_R\sum_{(i,j,r)\in\mathcal E_D}
  \sum_{k,\ell\in\mathcal V}Y_{ik}Y_{j\ell}P_{ij,k\ell},
  \label{eq:structured_assignment}
\end{aligned}
\end{equation}
subject to
\begin{equation}
  \sum_{k\in\mathcal V\cup\{\varnothing_i\}}Y_{ik}=1
  \quad \forall i,
  \qquad
  \sum_{i:s_i=s}\eta_iY_{ik}\leq1
  \quad \forall k,s.
  \label{eq:assignment_constraints}
\end{equation}
The first constraint gives each entity a real or null interpretation. The second prevents simultaneous actions from producing conflicting targets on the same part, while allowing reference-only entities to share a landmark. Because articulated assets and commands have few nodes, the quadratic assignment can be solved exactly as a mixed-integer linear program after standard product linearization.

\textbf{Structured confidence and abstention.}
For an accepted match $k_i^*$, let the best valid global interpretation that forbids this match define the structured margin
\begin{equation}
  \Delta_i=
  \mathcal E(\mathbf Y^*)-
  \max_{\mathbf Y\in\mathcal Y:\,Y_{ik_i^*}=0}
  \mathcal E(\mathbf Y).
  \label{eq:structured_margin}
\end{equation}
A directive is executed only when it is assigned to a real part and $\Delta_i\geq\gamma$, where $\gamma$ is calibrated on validation data. Otherwise, the system returns competing candidates or requests a qualifier rather than resolving ambiguity by part index.
Abstention produces no motion for that directive and returns either competing candidates or a request for a qualifier. It never falls back to a default part or silently clips magnitude.

\subsection{Calibrated Language-to-Joint Actuation}
\label{sec:actuation}

This stage is shown in panel~(D) of Figure~\ref{fig:arch}, where accepted directives become typed joint targets that are posed by forward kinematics.
\paragraph{Endpoint semantics.}
The order of the two observations does not determine which state is \emph{open}. When endpoint labels are unavailable, a two-way classifier uses part-centric renders $I_e^{(0)},I_e^{(1)}$ and the part anchor to estimate
\begin{equation}
  p_e^{\rm open}(s)=
  \operatorname{softmax}_{s\in\{0,1\}}
  g_{\rm end}\bigl(\phi_I(I_e^{(s)}),\mathbf a_k\bigr).
  \label{eq:endpoint_classifier}
\end{equation}
The selected state defines $q_e^{\rm open}$ and the other state defines $q_e^{\rm closed}$. Open/close commands abstain when the probability margin is below a validation threshold. These values are observed semantic anchors, not claims about the complete mechanical range.

\paragraph{Typed target conversion.}
The parser distinguishes endpoint, absolute-fraction, relative-fraction, absolute-metric, and relative-metric requests. For selected edge $e$, let $q_t$ be the endpoint named by $v_i$, $q_o$ the opposite endpoint, $q_c=q_e^{\rm cur}$, $\phi_i\in[0,1]$ a fraction, and $\delta_i'$ a metric amount converted to radians or meters. With $s_v=\operatorname{sign}(q_t-q_o)$, the requested target is
\begin{equation}
  \widetilde q_e^*=
  \begin{cases}
    q_t, & \text{endpoint},\\
    (1-\phi_i)q_o+\phi_iq_t,
      & \text{absolute fraction},\\
    q_c+\phi_i(q_t-q_c),
      & \text{relative fraction},\\
    q_o+s_v\delta_i',
      & \text{absolute metric},\\
    q_c+s_v\delta_i',
      & \text{relative metric}.
  \end{cases}
  \label{eq:typed_target}
\end{equation}
This explicitly distinguishes \emph{halfway open}, \emph{open halfway from the current state}, \emph{open by $20^\circ$}, and \emph{open to $20^\circ$}. A target is accepted only if its unit is compatible with the selected joint and $\widetilde q_e^*\in\mathcal I_e$. Otherwise, the directive is rejected rather than silently clipped to a different action.

For an accepted target $q_e^*=\widetilde q_e^*$, a rendered clip eases from the current configuration to the target with the cubic \emph{timing} curve $h(t)=3t^2-2t^3$, $h(0)=0$, $h(1)=1$:
\begin{equation}
  q_e(t)=q_e^{\rm cur}+h(t)\,(q_e^*-q_e^{\rm cur}),
  \qquad t\in[0,1],
  \label{eq:smooth_trajectory}
\end{equation}
with unselected joints held fixed. Here $t$ is normalised \emph{time}: $h$ only shapes the velocity profile of the animation, is monotonic on $[0,1]$, and is never evaluated outside it. When the accepted target is the opposite endpoint and the current configuration is the rest state, this clip coincides with the observed state $0\!\to\!1$ actuation; for a fractional or metric target it instead eases to the requested intermediate configuration, so current-to-target is the full $0\!\to\!1$ actuation only in that endpoint case. Substituting $\mathbf q(t)$ into Eq.~\eqref{eq:forward_kinematics} produces a graph-consistent motion clip.

Extrapolation past the observed states is a \emph{separate} construction and does not reuse the timing curve. Because the joint axis and pivot are recovered geometrically rather than clipped to the two observed states, we can scale the recovered state-to-state transform by a signed factor $s$ about the rest state,
\begin{equation}
  q_e(s)=q_e^{(0)}+s\,\bigl(q_e^{(1)}-q_e^{(0)}\bigr),
  \qquad s\in[-1,1],
  \label{eq:linear_fraction}
\end{equation}
where $q_e^{(0)}$ is the observed rest state (state $0$, the reconstructed pose) and $q_e^{(1)}$ the second observed state, in observation order and with no open/closed semantics attached. Thus $s=0$ reproduces state $0$, $s=1$ reproduces state $1$, $s\in[0,1]$ interpolates between the two observed states, and $s<0$ continues the motion linearly in reverse past the rest state; each joint coordinate (a revolute angle or a prismatic offset) is scaled linearly by $s$. This is the variable swept for the \textsc{L-M} metric and the actuation figures, and it is distinct from the animation timing $t$: the two share only the values at the observed states. Unlike a commanded target, which is accepted only when $\widetilde q_e^*\in\mathcal I_e$, this sweep is deliberately \emph{not} clipped to the recovered interval $\mathcal I_e$: for $s<0$ the coordinate $q_e(s)=2q_e^{(0)}-q_e^{(1)}$ at $s=-1$ can lie outside $\mathcal I_e$, and we intend it to, because the negative-$s$ poses probe the learned transform continued past the observed range rather than a feasible actuation. Joint-limit feasibility (whether $q_e(s)\in\mathcal I_e$) is thus not asserted for the sweep, and it is separate from collision and contact feasibility, which are downstream planning constraints and are likewise not implied by kinematic validity.

\subsection{Learning and Inference}
\label{sec:learning}

The feature field is optimized separately for each asset using Eq.~\eqref{eq:field_distillation}. The command resolver is trained across assets from ground-truth directive-part assignments, relations, null examples, and endpoint labels. With structured energy $\mathcal E_\theta(\mathbf Y)$ from Eq.~\eqref{eq:structured_assignment}, the directive, relation, and null assignments are supervised by the structured max-margin loss
\begin{equation}
  \mathcal L_{\rm str}
  =\max_{\mathbf Y\in\mathcal Y}
  \bigl[\mathcal E_\theta(\mathbf Y)
  +\ell(\mathbf Y,\mathbf Y^{\rm gt})\bigr]
  -\mathcal E_\theta(\mathbf Y^{\rm gt}),
  \label{eq:training_objective}
\end{equation}
where $\ell$ penalizes incorrect real, null, and relation assignments. The endpoint classifier is trained separately with a two-way cross-entropy over the endpoint labels of the two observed states. Training includes repeated parts, missing and spurious proposals, paraphrases, ambiguous commands, and incompatible actions; object and composition splits are disjoint.

At inference, \artlang{} builds $(\mathcal A,\Phi)$ offline, performs per-view proposal binding, parses the command, solves Eq.~\eqref{eq:structured_assignment}, rejects uncertain or infeasible assignments, and executes the remaining targets through the typed target conversion in Eq.~\eqref{eq:typed_target} and the forward-kinematics Eq.~\eqref{eq:forward_kinematics}.

\subsection{Training Details}
\label{sec:supp_training}
We describe the full training setup. The feature field is optimized separately for each asset. The cross-asset resolver is trained with two objectives: the structured max-margin loss above for the directive, relation, and null assignments, and a two-way cross-entropy on the endpoint labels for its endpoint classifier. The articulation graph itself is produced by the reconstruction backend and is not learned by us.

\paragraph{Feature field.}
For each reconstructed asset we distill a dense open-vocabulary feature field onto the canonical surface using Eq.~\eqref{eq:field_distillation}. Multi-view frames are encoded by a frozen FG-CLIP2~\cite{fgclip2} vision-language encoder, and the field is rendered through the same differentiable rasterizer that produces the geometry so that the rendered features match the encoder features under the cosine objective. The field is a per-point head on the reconstructed surface, optimized with Adam at a learning rate of $1\times10^{-3}$ until the distillation loss plateaus, typically a few thousand iterations per asset. The visibility- and area-weighted part prototypes are then aggregated in a single forward pass over the surface and cached, so that at query time only the lightweight part-level descriptor and the retained dense field are needed.

\paragraph{Command resolver.}
The structured resolver that produces the directive-to-part assignment is trained once across assets, not per object. Supervision comes from ground-truth directive-part assignments, relation labels, explicit reference frames, and null examples. We optimize the structured max-margin objective of Eq.~\eqref{eq:training_objective}, where the loss-augmented inference is solved exactly as a small mixed-integer program because each asset and command contains few nodes. The relation scorers and the unary semantic, spatial, and action heads are trained jointly. Training commands are generated procedurally to include repeated parts, missing and spurious proposals, paraphrases of the same instruction, reference-only entities, ambiguous commands with more than one valid target, and actions that are incompatible with the selected joint, so that the null node and the abstention margin receive gradient. Object and composition splits are disjoint between training and evaluation.

\paragraph{Endpoint classifier.}
The two-way endpoint classifier that decides which observed state is \emph{open} is trained from part-centric renders of the two states together with the part anchor, using the endpoint labels available in the synthetic data. Open/close commands abstain when its probability margin $|p^{\rm open}(0)-p^{\rm open}(1)|$ falls below a fixed threshold of $0.1$, chosen on held-out validation assets.

\paragraph{Implementation.}
The reconstruction backend is instantiated with ArtMesh~\cite{artmesh}. Text embeddings for anchors and queries use the FG-CLIP2 text encoder that matches the distilled image features. The two abstention thresholds, the structured margin $\gamma=0.005$ and the endpoint margin $0.1$, are selected on validation data and held fixed across every experiment. Exact per-stage iteration counts and the hardware configuration are reported in the released configuration files.

\subsection{Hyperparameters, Abstention Gate, and Solver}
\label{sec:supp-hparams}
The semantic score of directive $i$ against part $k$ mixes an anchor cosine with an area-weighted soft-max pooling of the dense distilled field, $s^{\text{sem}}_{ik}=\beta\langle t_i,a_k\rangle+(1-\beta)\tau_f\log(\tfrac{1}{A_k}\int_{G_k}e^{\langle t_i,f_\theta(x)\rangle/\tau_f}dA)$, and the unary energy is $U_{ik}=\lambda_{\text{sem}}s^{\text{sem}}_{ik}+\lambda_{\text{sp}}s^{\text{sp}}_{ik}+\lambda_{\text{act}}s^{\text{act}}_{ik}$, with an incompatible action forcing $U_{ik}=-\infty$.
The null score is $\nu_i=b_{\text{null}}+w_{\text{ent}}H(\mathrm{softmax}(s^{\text{sem}}_i/\tau_s))+w_{\text{conf}}(1-c_i)$, where $c_i\in[0,1]$ is the parser confidence for directive $i$; an under-specified directive whose semantic scores are diffuse is thus drawn toward its null node, a directive executes only when the structured margin satisfies $\Delta_i\ge\gamma$, and otherwise abstains.
The pairwise relation term is a linear scorer $g_r$ over an oriented $6$-D geometric feature, where the orientation signs the geometry per relation so that ``relation satisfied'' is always the positive direction, letting one shared weight generalise across polarity.
The global assignment is an exact product-linearised MILP solved with PuLP's CBC backend at its default optimality tolerance, with one binary per (directive, part) plus a private null node, an exactly-one constraint per directive, a per-part per-stage capacity constraint for actionable directives, and a McCormick linearisation of each relation edge.
Table~\ref{tab:supp-hparams} lists the fixed scoring, abstention, and binding hyperparameters; the energy weights are trained on the reported dataset, giving $\lambda=(1.63,0.23,0.53)$, $\lambda_R=1.0$, $g_r=(0.08,0.52,0,0,0,0)$, $b_{\text{null}}=0.43$, with the calibrated abstention margin $\gamma=0.005$.

\begin{table}[htbp]
\centering\small
\begin{tabular}{ll}
\toprule
Symbol & Value \\
\midrule
$\beta$ (anchor / dense-field mix)        & $0.6$ \\
$\tau_f$ (dense-field pooling temp.)      & $0.1$ \\
$\tau_s$ (semantic soft-max temp.)        & $0.1$ \\
$w_{\text{ent}}$ (null-score entropy weight) & $0.25$ \\
$w_{\text{conf}}$ (null-score parser-confidence weight) & $1.0$ \\
$\tau_{\text{dummy}}$ (private-dummy cost)& $0.5$ \\
\bottomrule
\end{tabular}
\caption{Fixed scoring, abstention, and binding hyperparameters. The learned energy weights ($\lambda$, $g_r$, $b_{\text{null}}$) and the calibrated abstention margin ($\gamma$) are reported in the text.}
\label{tab:supp-hparams}
\end{table}

\subsection{Action-Joint Compatibility}
\label{sec:supp-actcompat}
The action channel applies a compatibility mask: a directive verb is admissible on a part only if the part is movable and its joint type matches the verb's allowed set, and a fixed part or a verb/type mismatch sets $s^{\text{act}}=-\infty$, so an incompatible pairing is pruned before the assignment. Every action verb the grammar emits reduces to \emph{open} or \emph{close}, which are admissible on both revolute and prismatic joints, and a directive whose verb names no endpoint or whose target is infeasible is rejected at target conversion (Eq.~\eqref{eq:typed_target}) rather than actuated.

\section{Experiments}
\label{sec:exp}

We evaluate \artlang{} on Articulate-100, built from PartNet-Mobility\cite{sapien}, in two settings: reconstruction from multi-view observations and mesh input, which removes geometry recovery. We first verify articulation quality, then evaluate semantic, spatial, and motion grounding through continuous actuation and controlled ablations. We also test on real SplArt captures.

\begin{table}[t]
\centering
\caption{\textbf{Mesh-input ablation.} Each variant removes one language competency while geometry is fixed. \textbf{Best}, \underline{Second}.}
\label{tab:ablation}
\setlength{\tabcolsep}{4pt}
\renewcommand{\arraystretch}{1.1}
\resizebox{0.8\linewidth}{!}{
\begin{tabular}{lcccc}
\toprule
 & Ours & NaiveCLIP & NoSpatial & NaiveWarp \\
\midrule
Axis Ang $\downarrow$ & \textbf{23.79} & -- & -- & \underline{26.06} \\
\midrule
Axis Pos $\downarrow$ & \textbf{0.08} & -- & -- & \underline{1.03} \\
\midrule
Part Motion $\downarrow$ & \textbf{6.24} & -- & -- & \underline{13.76} \\
\midrule
CD-m $\downarrow$ & \textbf{26.78} & -- & -- & \underline{89.42} \\
\midrule
L-M Alignment $\downarrow$ & \textbf{70.42} & 392.52 & \underline{89.32} & 127.38 \\
\midrule
L-Sem Alignment $\uparrow$ & \textbf{0.78} & 0.33 & -- & \underline{0.71} \\
\midrule
L-Spat Alignment $\uparrow$ & \textbf{0.61} & 0.08 & 0.03 & \underline{0.50} \\
\bottomrule
\end{tabular}
}
\end{table}

\begin{figure}[t]
\centering
\includegraphics[width=\linewidth]{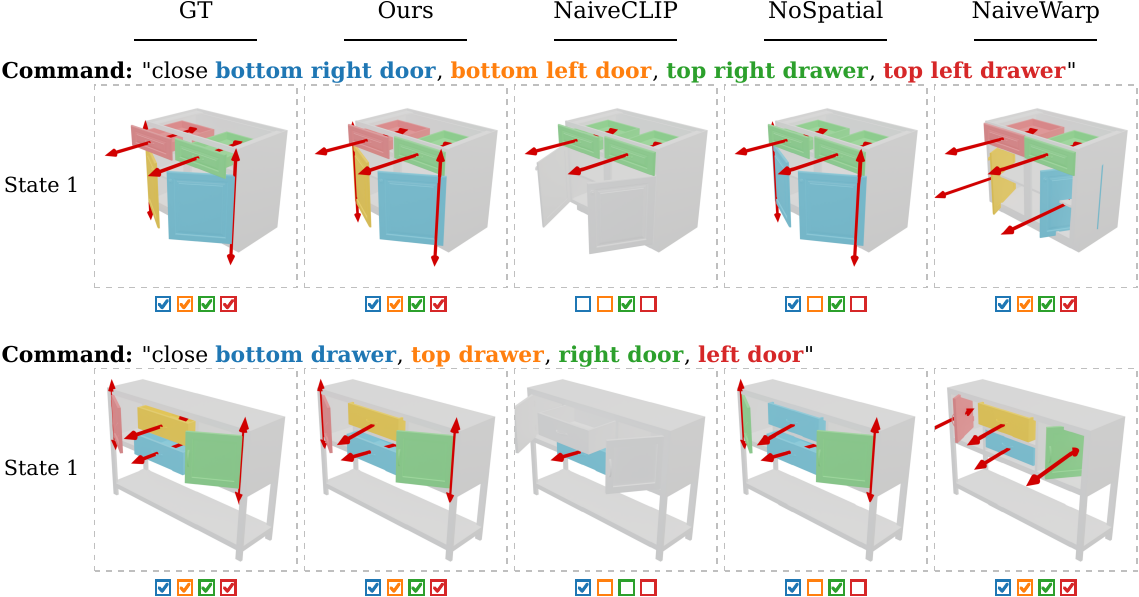}
\caption{\textbf{Language grounding ablation.} NaiveCLIP~\cite{clip} weakens semantic assignment, NoSpatial removes spatial disambiguation, and NaiveWarp replaces per-part articulated motion with a centroid-based warp. Same-colored movable parts indicate those selected by each method for the command.}
\label{fig:ablation}
\end{figure}

\subsection{Protocol}
\label{sec:exp_protocol}

Axis Ang and Axis Pos measure joint-axis error, Part Motion measures articulated end-state error, and CD-s and CD-m are Chamfer distances for the static, movable, and complete object. Language performance is measured by L-Sem for noun-to-part grounding, L-Spat for spatial disambiguation, and L-M for commanded motion trajectories. Kinematic, Chamfer, and motion metrics are errors, so lower is better; L-Sem and L-Spat are overlaps, so higher is better.

We compare with OPD~\cite{opd} and Articulate-Anything~\cite{articulate-anything} for articulation, and ArtGS~\cite{artgs} and GaussianArt~\cite{gaussianart} for reconstruction. NaiveCLIP assigns part names using raw CLIP similarity without projection-based assignment or explicit spatial grounding.

\subsection{Language-Alignment Metrics}
\label{sec:supp-metrics}
We report three complementary language-alignment metrics, each computed per object and then macro-averaged over the evaluation set. \textsc{L-Sem} (LSE-A) and \textsc{L-Spat} (LSP-A) are name-bucket mask intersection-over-union scores: every predicted part is coloured by the noun it was bound to and every ground-truth part by its annotated noun, each name is split into a noun key and a spatial-modifier key (any token in a fixed spatial-word list of \texttt{left}, \texttt{right}, \texttt{top}, \texttt{bottom}, \texttt{upper}, \texttt{lower}, \texttt{front}, \texttt{back}, \texttt{middle}, \texttt{center}, \texttt{inner}, \texttt{outer}, the superlative \texttt{-most} forms, ordinals, and digit indices is a modifier, and every remaining token is the noun), and for each held-out view we render the predicted part segmentation, assign each pixel to the $\arg\max$ per-part coverage field (pixels with total coverage $\le 0.5$ are left unassigned), and group both predicted and ground-truth masks into buckets.
\textsc{L-Sem} is the mean IoU over all (view, ground-truth noun-bucket) pairs and \textsc{L-Spat} the mean IoU over the ground-truth spatial-modifier buckets, both in $[0,1]$ and higher-is-better, so they measure whether the noun words and the directional words respectively land on the same regions; the buckets are keyed by name rather than by part index, a ground-truth key absent from the prediction is scored against an empty predicted mask (IoU $0$), and predicted keys absent from the ground truth are ignored.
\textsc{L-M} (LM-A) is the language motion-alignment metric: it measures whether the motion produced by a language-grounded command traces the correct trajectory. Ground-truth joints are paired to predicted parts by a Hungarian assignment on name agreement (exact match preferred, shared head word next, disjoint names last), and the paired parts are scored by trajectory Chamfer; naming quality itself is measured by \textsc{L-Sem} and \textsc{L-Spat}. The paired parts are swept \emph{linearly} along their joints at five fractions $s\in\{-1,-0.5,0,0.5,1\}$ of the recovered state-to-state transform (each joint coordinate scaled linearly by $s$, Eq.~\eqref{eq:linear_fraction}), and at each fraction $10{,}000$ surface samples per part give a symmetric Chamfer distance, where $\overline{d_{A\to B}}$ is the mean nearest-neighbour \emph{Euclidean} distance (not squared) from points of $A$ to $B$,
\begin{equation}
\mathrm{CD}(A,B)=\tfrac{1}{2}\!\left(\overline{d_{A\to B}}+\overline{d_{B\to A}}\right)\times\bigl(10^{3}\,\mathrm{mm/m}\bigr).
\end{equation}
\textsc{L-M} is the mean of these per-fraction Chamfer distances in millimetres, lower-is-better; a value such as $\textsc{L-M}=70.42$ denotes a $70.42$\,mm average surface deviation of the moving part across its swept trajectory. A geometry-paired counterpart, which matches joints to parts by geometry instead of names, provides a name-independent motion score.
All three metrics are averaged within an object then macro-averaged across objects. A ground-truth bucket with no predicted pixels receives an IoU of $0$; a part left unnamed at binding time otherwise removes its pixels from the corresponding predicted bucket, generally reducing but not necessarily zeroing that bucket's IoU when same-name siblings remain, so abstentions are penalised rather than excluded.
The directive-level abstention protocol (accuracy with and without abstention on an ambiguity stress subset) is reported separately in the supplementary material.

\subsection{Language Grounding Ablation}
\label{sec:exp_ablation}

Table~\ref{tab:ablation} and Figure~\ref{fig:ablation} are our main evaluations of language-conditioned control. One component is removed at a time, isolating semantic grounding, spatial disambiguation, and motion execution from reconstruction quality.

\textbf{Semantic grounding.}
NaiveCLIP replaces projection-based semantic-geometric assignment with raw CLIP similarity. L-Sem drops substantially, showing that dense similarity alone is insufficient for binding language to persistent parts. L-Spat also decreases because spatial reasoning cannot recover the intended instance after selecting the wrong semantic candidate.

\textbf{Spatial disambiguation.}
NoSpatial removes explicit-frame spatial evidence while retaining semantic naming.L-Spat drops sharply for repeated parts such as doors and drawers. This shows that modifiers including \emph{left}, \emph{right}, \emph{top}, and \emph{below} require an explicit spatial grounding component.

\textbf{Motion grounding.}
NaiveWarp preserves semantic and spatial selection but replaces per-part joints with a single warp about the object centroid. L-Sem and L-Spat remain comparatively stable, while joint error, Part Motion, and L-M degrade substantially. Correct part identification must therefore be paired with a rigid, executable part trajectory.

Each ablation primarily harms the axis it removes, supporting the decomposition of \artlang{} into semantic identification, explicit-frame spatial resolution, and articulated motion execution. Reporting these axes separately exposes failures that a single aggregate score would hide.

\subsection{Language-Driven Continuous Actuation}
\label{sec:exp_continuous}

\begin{figure*}[t]
\centering
\includegraphics[width=\linewidth]{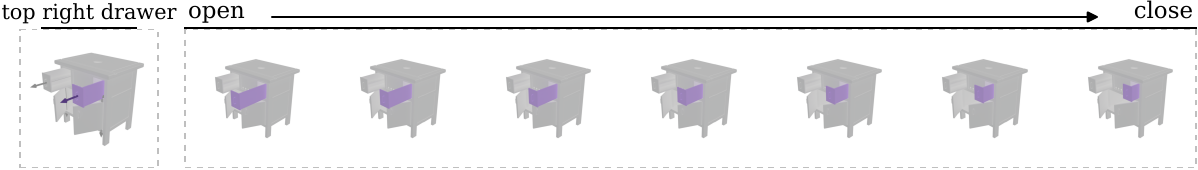}
\caption{\textbf{Continuous language-driven actuation.} The grounded part moves along its recovered joint as $t$ varies from $-1$ to $1$, while all other parts remain fixed.}
\label{fig:text_move}
\end{figure*}

A grounded directive selects a persistent part and places it at a continuous coordinate of its recovered joint. Figure~\ref{fig:text_move} sweeps the actuation parameter $t$ while keeping all other parts fixed: $t=0$ is the rest state, $t=1$ is the learned end state, intermediate values produce unseen poses, and negative values extrapolate in the opposite direction.

The same coordinate supports modifiers such as \emph{halfway} and explicit angular or translational targets. Because the sweep follows the recovered joint rather than a ground-truth trajectory, it also reveals errors in the estimated axis and motion range.

\subsection{Articulation Quality}
\label{sec:exp_articulation}

\textbf{Reconstruction.}
\definecolor{lightblue}{RGB}{225, 232, 240}

\begin{table}[t]
\centering
\caption{\textbf{Reconstruction on Articulate-100.} \artlang{} matches ArtMesh in joint and geometry accuracy, and both outperform Gaussian baselines. Lower is better.}
\label{tab:recon}
\setlength{\tabcolsep}{4pt}
\renewcommand{\arraystretch}{1.1}
\resizebox{0.7\linewidth}{!}{
\begin{tabular}{lcccc}
\toprule
 & ArtMesh & Ours & ArtGS & GaussianArt \\
\midrule
Axis Ang $\downarrow$ & \textbf{3.48} & \underline{3.62} & 10.53 & 34.45 \\
\midrule
Axis Pos $\downarrow$ & \underline{0.03} & \textbf{0.01} & 0.11 & 0.33 \\
\midrule
Part Motion $\downarrow$ & 3.63 & \textbf{1.69} & \underline{3.50} & 16.48 \\
\midrule
CD-s $\downarrow$ & \textbf{18.98} & \underline{19.24} & 23.56 & 24.63 \\
\midrule
CD-m $\downarrow$ & \textbf{15.01} & \underline{16.62} & 151.41 & 38.43 \\
\bottomrule
\end{tabular}
}
\end{table}

\begin{figure}[t]
\centering
\includegraphics[width=\linewidth]{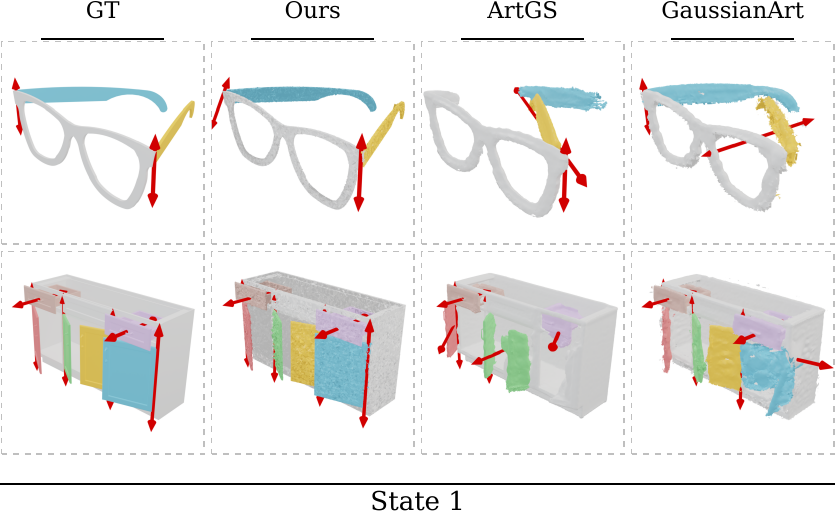}
\caption{\textbf{Reconstruction on Articulate-100.} Each row shows the ground truth, our reconstruction, and the baselines, together with part segmentation and recovered motion.
Full two-state results are provided in Figure~\ref{fig:supp_recon}.
}
\label{fig:recon}
\end{figure}

Table~\ref{tab:recon} and Figure~\ref{fig:recon} evaluates the complete reconstruction pipeline. \artlang{} matches the ArtMesh backbone in geometry and joint recovery, showing that the language field does not reduce reconstruction quality. Both outperform the Gaussian baselines on geometry and joint estimation.

\textbf{Mesh input.}
\begin{table}[t]
\centering
\caption{\textbf{Mesh-input results on Articulate-100.} We compare three \artlang{} input regimes with OPD and Articulate-Anything. Errors are lower-is-better; semantic and spatial overlaps are higher-is-better.}
\label{tab:mesh}
\setlength{\tabcolsep}{4pt}
\renewcommand{\arraystretch}{1.1}
\resizebox{\linewidth}{!}{
\begin{tabular}{lccccc}
\toprule
 & OPD & Art-Anything & Ours-Mesh & Ours-MV & Ours-SV \\
\midrule
Axis Ang $\downarrow$ & 34.88 & 26.57 & \textbf{6.98} & \underline{11.29} & 23.79 \\
\midrule
Axis Pos $\downarrow$ & 4.53 & 2.08 & \textbf{0.08} & \underline{0.09} & \textbf{0.08} \\
\midrule
Part Motion $\downarrow$ & 25.61 & 17.81 & \textbf{4.25} & \underline{4.89} & 6.24 \\
\midrule
CD-m $\downarrow$ & 240.77 & 114.92 & \textbf{10.26} & \underline{14.69} & 26.78 \\
\midrule
L-M Alignment $\downarrow$ & -- & -- & \textbf{50.71} & \underline{57.52} & 70.42 \\
\midrule
L-Sem Alignment $\uparrow$ & -- & -- & \textbf{0.86} & \underline{0.81} & 0.78 \\
\midrule
L-Spat Alignment $\uparrow$ & -- & -- & \textbf{0.71} & \underline{0.63} & 0.61 \\
\bottomrule
\end{tabular}
}
\end{table}

\begin{figure}[t]
\centering
\includegraphics[width=\linewidth]{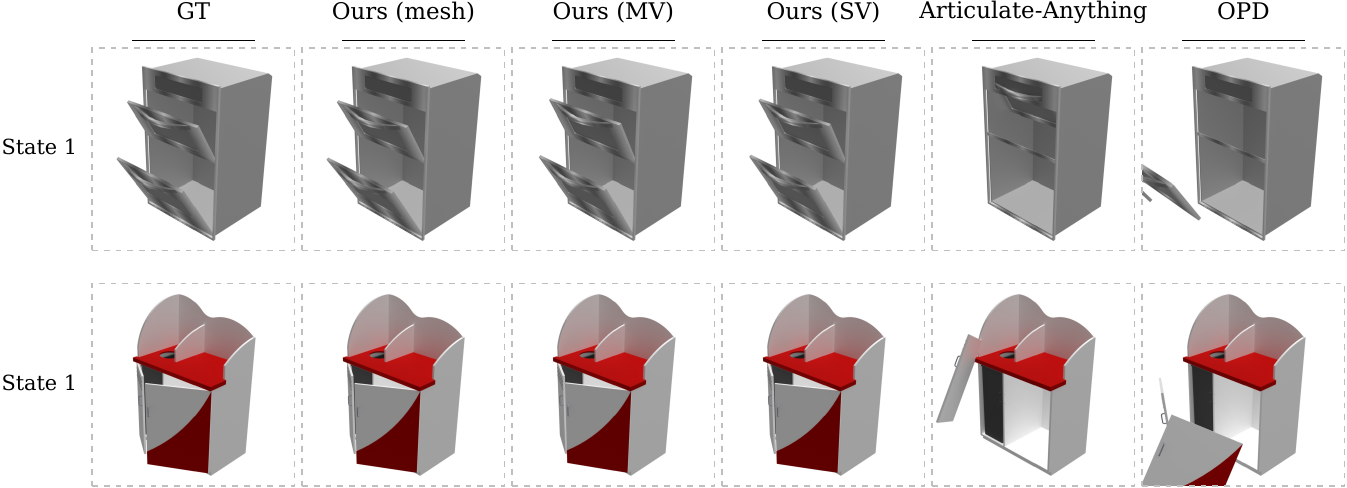}
\caption{\textbf{Mesh-input articulation results.} Columns show the ground truth, our three motion-observation settings, Articulate-Anything, and OPD.}
\label{fig:meshinput}
\end{figure}

Table~\ref{tab:mesh} and Figure~\ref{fig:meshinput} evaluates articulation when the canonical mesh is given. OPD receives up to eight RGB-D views and aggregates per-view detections, while Articulate-Anything receives a part-segmented mesh rendered at several actuated states for joint inference by its vision-language model.

Our variants recover substantially more accurate articulation than both baselines, with performance degrading gradually as motion evidence decreases from the full end mesh to multi-view and single-view observations. These results establish that the recovered joints are sufficiently accurate for language-driven actuation.

\subsection{Real Captures}
\label{sec:exp_splart}

\begin{figure}[t]
\centering
\includegraphics[width=\linewidth]{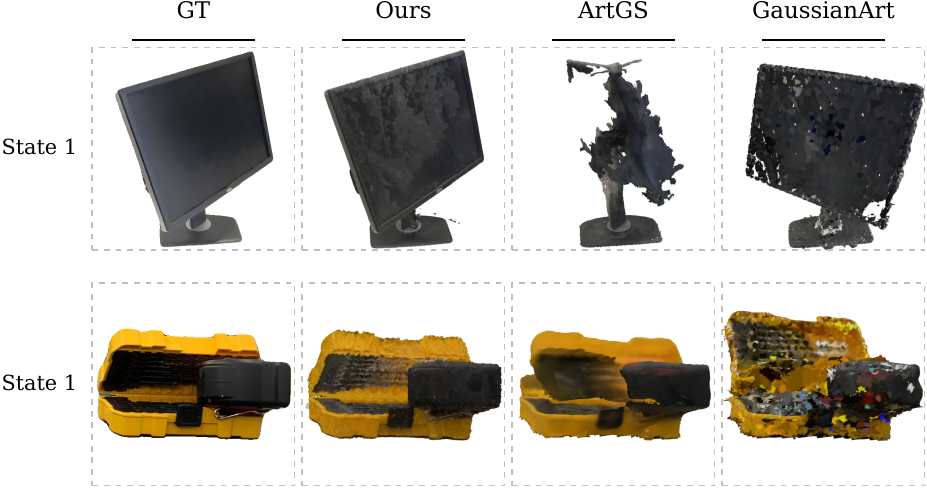}
\caption{\textbf{Real SplArt captures.} The language handle grounds and actuates parts despite noisy geometry and ambiguous boundaries. SplArt provides neither ground-truth meshes nor the original capture lighting, so rendered appearance may differ from the reference images.}
\label{fig:splart}
\end{figure}

Figure~\ref{fig:splart} applies the same pipeline to real SplArt~\cite{splart} objects. Despite noisier geometry and less distinct boundaries, semantic and spatial grounding identify the intended part, which is then actuated along its recovered joint. Users can therefore control captured objects through descriptions rather than part indices or numerical joint parameters.

\section{Conclusion}
\label{sec:conclusion}

We presented \artlang{}, which makes reconstructed articulated objects language-addressable by grounding semantic parts, spatial instances, and motion commands to persistent parts and recovered joints. Experiments on Articulate-100, mesh input, and real SplArt captures show effective semantic, spatial, and motion control while preserving articulation quality.

Several limitations remain. The current pipeline requires two observed articulation states, assumes a meaningful canonical frame for directional language, and supports a limited motion vocabulary without contact reasoning. Future work could extend the language handle to single-state reconstruction, richer spatial frames, contact-aware manipulation, physical constraints, and a broader range of actions. The factorized semantic, spatial, and motion structure provides a natural foundation for these extensions.

\clearpage
{
    \small
    \bibliographystyle{ieeenat_fullname}
    \bibliography{main}

@String(CVPR= {Proceedings of the IEEE/CVF Conference on Computer Vision and Pattern Recognition})

@String(ICCV= {Int. Conf. Comput. Vis.})

@String(ECCV= {Eur. Conf. Comput. Vis.})

@String(CVPR  = {CVPR})

@String(ICCV  = {ICCV})

@String(ECCV  = {ECCV})

@misc{gaussianart,
      title={GaussianArt: Unified Modeling of Geometry and Motion for Articulated Objects}, 
      author={Licheng Shen and Saining Zhang and Honghan Li and Peilin Yang and Zihao Huang and Zongzheng Zhang and Hao Zhao},
      year={2025},
      eprint={2508.14891},
      archivePrefix={arXiv},
      primaryClass={cs.CV},
      url={https://arxiv.org/abs/2508.14891}, 
}

@misc{meshsplatting,
      title={MeshSplatting: Differentiable Rendering with Opaque Meshes}, 
      author={Jan Held and Sanghyun Son and Renaud Vandeghen and Daniel Rebain and Matheus Gadelha and Yi Zhou and Anthony Cioppa and Ming C. Lin and Marc Van Droogenbroeck and Andrea Tagliasacchi},
      year={2025},
      eprint={2512.06818},
      archivePrefix={arXiv},
      primaryClass={cs.CV},
      url={https://arxiv.org/abs/2512.06818}, 
}

@misc{3dgs,
      title={3D Gaussian Splatting for Real-Time Radiance Field Rendering}, 
      author={Bernhard Kerbl and Georgios Kopanas and Thomas Leimkühler and George Drettakis},
      year={2023},
      eprint={2308.04079},
      archivePrefix={arXiv},
      primaryClass={cs.GR},
      url={https://arxiv.org/abs/2308.04079}, 
}

@inproceedings{artgs,
  title={Building Interactable Replicas of Complex Articulated Objects via Gaussian Splatting},
  author={Liu, Yu and Jia, Baoxiong and Lu, Ruijie and Ni, Junfeng and Zhu, Song-Chun and Huang, Siyuan},
  booktitle={The Thirteenth International Conference on Learning Representations},
  year={2025},
}

@inproceedings{paris,
  title={PARIS: Part-level Reconstruction and Motion Analysis for Articulated Objects},
  author={Liu, Jiayi and Mahdavi-Amiri, Ali and Savva, Manolis},
  booktitle={Proceedings of the IEEE/CVF International Conference on Computer Vision},
  pages={352--363},
  year={2023}
}

@inproceedings{cla-nerf,
      title={CLA-NeRF: Category-Level Articulated Neural Radiance Field}, 
      author={Wei-Cheng Tseng and Hung-Ju Liao and Yen-Chen Lin and Min Sun},
      booktitle={ICRA},
      year={2022},
}

@article{asdf,
  author    = {Jiteng Mu and
               Weichao Qiu and
               Adam Kortylewski and
               Alan L. Yuille and
               Nuno Vasconcelos and
               Xiaolong Wang},
  title     = {{A-SDF:} Learning Disentangled Signed Distance Functions for Articulated
               Shape Representation},
  booktitle = {ICCV},
  pages = {12981--12991},
  year      = {2021},
}

@misc{splart,
    title={SplArt: Articulation Estimation and Part-Level Reconstruction with 3D Gaussian Splatting}, 
    author={Shengjie Lin and Jiading Fang and Muhammad Zubair Irshad and Vitor Campagnolo Guizilini and Rares Andrei Ambrus and Greg Shakhnarovich and Matthew R. Walter},
    year={2025},
    eprint={2506.03594},
    archivePrefix={arXiv},
    primaryClass={cs.GR},
    url={https://arxiv.org/abs/2506.03594}, 
}

@misc{part2gs,
      title={Part$^{2}$GS: Part-aware Modeling of Articulated Objects using 3D Gaussian Splatting}, 
      author={Tianjiao Yu and Vedant Shah and Muntasir Wahed and Ying Shen and Kiet A. Nguyen and Ismini Lourentzou},
      year={2026},
      eprint={2506.17212},
      archivePrefix={arXiv},
      primaryClass={cs.CV},
      url={https://arxiv.org/abs/2506.17212}, 
}

@inproceedings{nerf,
  title={NeRF: Representing Scenes as Neural Radiance Fields for View Synthesis},
  author={Ben Mildenhall and Pratul P. Srinivasan and Matthew Tancik and Jonathan T. Barron and Ravi Ramamoorthi and Ren Ng},
  year={2020},
  booktitle={ECCV},
}

@article{neus,
  title={NeuS: Learning Neural Implicit Surfaces by Volume Rendering for Multi-view Reconstruction},
  author={Wang, Peng and Liu, Lingjie and Liu, Yuan and Theobalt, Christian and Komura, Taku and Wang, Wenping},
  journal={arXiv preprint arXiv:2106.10689},
  year={2021}
}

@article{articulate-nerf,
  title={Articulate your NeRF: Unsupervised articulated object modeling via conditional view synthesis},
  author={Deng, Jianning and Subr, Kartic and Bilen, Hakan},
  journal={arXiv preprint arXiv:2406.16623},
  year={2024}
}

@inproceedings{digitaltwinart,
    title={Neural Implicit Representation for Building Digital Twins of Unknown Articulated Objects}, 
    author={Yijia Weng and Bowen Wen and Jonathan Tremblay and Valts Blukis and Dieter Fox and Leonidas Guibas and Stan Birchfield},
    booktitle={CVPR},
    year={2024}
}

@misc{shape2motion,
      title={Shape2Motion: Joint Analysis of Motion Parts and Attributes from 3D Shapes}, 
      author={Xiaogang Wang and Bin Zhou and Yahao Shi and Xiaowu Chen and Qinping Zhao and Kai Xu},
      year={2019},
      eprint={1903.03911},
      archivePrefix={arXiv},
      primaryClass={cs.CV},
      url={https://arxiv.org/abs/1903.03911}, 
}

@inproceedings{captra,
	title={CAPTRA: CAtegory-level Pose Tracking for Rigid and Articulated Objects from Point Clouds},
	author={Weng, Yijia and Wang, He and Zhou, Qiang and Qin, Yuzhe and Duan, Yueqi and Fan, Qingnan and Chen, Baoquan and Su, Hao and Guibas, Leonidas J.},
	booktitle={Proceedings of the IEEE International Conference on Computer Vision (ICCV)},
    	month={October},
	year={2021},
    	pages={13209-13218}
}

@inProceedings{where2act,
    title={Where2Act: From Pixels to Actions for Articulated 3D Objects},
    author={Mo, Kaichun and Guibas, Leonidas and Mukadam, Mustafa and Gupta, Abhinav and Tulsiani, Shubham},
    year={2021},
    booktitle={International Conference on Computer Vision (ICCV)}
}

@misc{sage,
      title={SAGE: Bridging Semantic and Actionable Parts for GEneralizable Articulated-Object Manipulation under Language Instructions}, 
      author={Haoran Geng and Songlin Wei and Congyue Deng and Bokui Shen and He Wang and Leonidas Guibas},
      year={2023},
      eprint={2312.01307},
      archivePrefix={arXiv},
      primaryClass={cs.RO}
}

@misc{real2code,
      title={Real2Code: Reconstruct Articulated Objects via Code Generation}, 
      author={Zhao Mandi and Yijia Weng and Dominik Bauer and Shuran Song},
      year={2024},
      eprint={2406.08474},
      archivePrefix={arXiv},
      primaryClass={cs.CV},
      url={https://arxiv.org/abs/2406.08474}, 
}

@misc{urdformer,
      title={URDFormer: A Pipeline for Constructing Articulated Simulation Environments from Real-World Images}, 
      author={Zoey Chen and Aaron Walsman and Marius Memmel and Kaichun Mo and Alex Fang and Karthikeya Vemuri and Alan Wu and Dieter Fox and Abhishek Gupta},
      year={2024},
      eprint={2405.11656},
      archivePrefix={arXiv},
      primaryClass={cs.RO},
      url={https://arxiv.org/abs/2405.11656}, 
}

@article{larm,
      title={LARM: A Large Articulated-Object Reconstruction Model}, 
      author={Yuan, Sylvia and Shi, Ruoxi and Wei, Xinyue and Zhang, Xiaoshuai and Su, Hao and Liu, Minghua},
      journal={arXiv preprint arXiv:2511.11563},
      year={2025},
}

@inproceedings{opd,
  title={OPD: Single-view 3D openable part detection},
  author={Jiang, Hanxiao and Mao, Yongsen and Savva, Manolis and Chang, Angel X},
  booktitle={Computer Vision--ECCV 2022: 17th European Conference, Tel Aviv, Israel, October 23--27, 2022, Proceedings, Part XXXIX},
  pages={410--426},
  year={2022},
  organization={Springer}
}

@article{opdmulti,
  title={OPDMulti: Openable Part Detection for Multiple Objects},
  author={Sun, Xiaohao and Jiang, Hanxiao and Savva, Manolis and Chang, Angel Xuan},
  journal={arXiv preprint arXiv:2303.14087},
  year={2023}
}

@article{singapo,
  title={{SINGAPO}: Single Image Controlled Generation of Articulated Parts in Object},
  author={Liu, Jiayi and Iliash, Denys and Chang, Angel X and Savva, Manolis and Mahdavi-Amiri, Ali},
  journal={arXiv preprint arXiv:2410.16499},
  year={2024}
}

@misc{dreamart,
      title={DreamArt: Generating Interactable Articulated Objects from a Single Image}, 
      author={Ruijie Lu and Yu Liu and Jiaxiang Tang and Junfeng Ni and Yuxiang Wang and Diwen Wan and Gang Zeng and Yixin Chen and Siyuan Huang},
      year={2025},
      eprint={2507.05763},
      archivePrefix={arXiv},
      primaryClass={cs.CV},
      url={https://arxiv.org/abs/2507.05763}, 
}

@misc{partrm,
      title={PartRM: Modeling Part-Level Dynamics with Large Cross-State Reconstruction Model}, 
      author={Mingju Gao and Yike Pan and Huan-ang Gao and Zongzheng Zhang and Wenyi Li and Hao Dong and Hao Tang and Li Yi and Hao Zhao},
      year={2025},
      eprint={2503.19913},
      archivePrefix={arXiv},
      primaryClass={cs.CV},
      url={https://arxiv.org/abs/2503.19913}, 
}

@article{articulate-anything,
  title={Articulate-Anything: Automatic Modeling of Articulated Objects via a Vision-Language Foundation Model},
  author={Le, Long and Xie, Jason and Liang, William and Wang, Hung-Ju and Yang, Yue and Ma, Yecheng Jason and Vedder, Kyle and Krishna, Arjun and Jayaraman, Dinesh and Eaton, Eric},
  journal={arXiv preprint arXiv:2410.13882},
  year={2024}
}

@article{a3vlm,
  title={A3VLM: Actionable Articulation-Aware Vision Language Model},
  author={Huang, Siyuan and Chang, Haonan and Liu, Yuhan and Zhu, Yimeng and Dong, Hao and Gao, Peng and Boularias, Abdeslam and Li, Hongsheng},
  journal={arXiv preprint arXiv:2406.07549},
  year={2024}
}

@misc{manipllm,
      title={ManipLLM: Embodied Multimodal Large Language Model for Object-Centric Robotic Manipulation}, 
      author={Xiaoqi Li and Mingxu Zhang and Yiran Geng and Haoran Geng and Yuxing Long and Yan Shen and Renrui Zhang and Jiaming Liu and Hao Dong},
      year={2023},
      eprint={2312.16217},
      archivePrefix={arXiv},
      primaryClass={cs.CV},
      url={https://arxiv.org/abs/2312.16217}, 
}

@misc{urdfanything,
      title={URDF-Anything: Constructing Articulated Objects with 3D Multimodal Language Model}, 
      author={Zhe Li and Xiang Bai and Jieyu Zhang and Zhuangzhe Wu and Che Xu and Ying Li and Chengkai Hou and Shanghang Zhang},
      year={2025},
      eprint={2511.00940},
      archivePrefix={arXiv},
      primaryClass={cs.RO},
      url={https://arxiv.org/abs/2511.00940}, 
}

@article{artmesh,
  title={ArtMesh: Part-Aware Articulated Mesh Fields with Motion-Consistent Dynamics},
  author={Yuan, Sylvia and Wang, Dan and Ramamoorthi, Ravi and Cui, Xinrui},
  journal={arXiv preprint arXiv:2605.16582},
  year={2026}
}

@article{articulated-gs,
  title={ArticulatedGS: Self-supervised Digital Twin Modeling of Articulated Objects using 3D Gaussian Splatting},
  author={Guo, Junfu and Xin, Yu and Liu, Gaoyi and Xu, Kai and Liu, Ligang and Hu, Ruizhen},
  journal={arXiv preprint arXiv:2503.08135},
  year={2025}
}

@article{reartgs,
  title={Reartgs: Reconstructing and generating articulated objects via 3d gaussian splatting with geometric and motion constraints},
  author={Wu, Di and Liu, Liu and Linli, Zhou and Huang, Anran and Song, Liangtu and Yu, Qiaojun and Wu, Qi and Lu, Cewu},
  journal={arXiv preprint arXiv:2503.06677},
  year={2025}
}

@inproceedings{2dgs,
    title={2D Gaussian Splatting for Geometrically Accurate Radiance Fields},
    author={Huang, Binbin and Yu, Zehao and Chen, Anpei and Geiger, Andreas and Gao, Shenghua},
    publisher = {Association for Computing Machinery},
    booktitle = {SIGGRAPH 2024 Conference Papers},
    year      = {2024},
    doi       = {10.1145/3641519.3657428}
}

@misc{rade-gs,
      title={RaDe-GS: Rasterizing Depth in Gaussian Splatting}, 
      author={Baowen Zhang and Chuan Fang and Rakesh Shrestha and Yixun Liang and Xiaoxiao Long and Ping Tan},
      year={2024},
      eprint={2406.01467},
      archivePrefix={arXiv},
      primaryClass={cs.GR},
      url={https://arxiv.org/abs/2406.01467}, 
}

@article{sugar,
        title={SuGaR: Surface-Aligned Gaussian Splatting for Efficient 3D Mesh Reconstruction and High-Quality Mesh Rendering},
        author={Gu{\'e}don, Antoine and Lepetit, Vincent},
        journal={CVPR},
        year={2024}
      }

@article{milo,
        author       = {Gu{\'e}don, Antoine and Gomez, Diego and Maruani, Nissim and Gong, Bingchen and Drettakis, George and Ovsjanikov, Maks},
        title        = {MILo: Mesh-In-the-Loop Gaussian Splatting for Detailed and Efficient Surface Reconstruction},
        journal      = {ACM Transactions on Graphics},
        number       = {},
        volume       = {},
        month        = {},
        year         = {2025},
        url          = {https://anttwo.github.io/milo/}
      }

@misc{triangle-splatting,
      title={Triangle Splatting for Real-Time Radiance Field Rendering}, 
      author={Jan Held and Renaud Vandeghen and Adrien Deliege and Abdullah Hamdi and Silvio Giancola and Anthony Cioppa and Andrea Vedaldi and Bernard Ghanem and Andrea Tagliasacchi and Marc Van Droogenbroeck},
      year={2025},
      eprint={2505.19175},
      archivePrefix={arXiv},
      primaryClass={cs.CV},
      url={https://arxiv.org/abs/2505.19175}, 
}

@InProceedings{sapien,
author = {Xiang, Fanbo and Qin, Yuzhe and Mo, Kaichun and Xia, Yikuan and Zhu, Hao and Liu, Fangchen and Liu, Minghua and Jiang, Hanxiao and Yuan, Yifu and Wang, He and Yi, Li and Chang, Angel X. and Guibas, Leonidas J. and Su, Hao},
title = {{SAPIEN}: A SimulAted Part-based Interactive ENvironment},
booktitle = {The IEEE Conference on Computer Vision and Pattern Recognition (CVPR)},
month = {June},
year = {2020}}

@inproceedings{meshart,
  title={Meshart: Generating articulated meshes with structure-guided transformers},
  author={Gao, Daoyi and Siddiqui, Yawar and Li, Lei and Dai, Angela},
  booktitle={Proceedings of the Computer Vision and Pattern Recognition Conference},
  pages={618--627},
  year={2025}
}

@inproceedings{artilatent,
  title={ArtiLatent: Realistic Articulated 3D Object Generation via Structured Latents},
  author={Chen, Honghua and Lan, Yushi and Chen, Yongwei and Pan, Xingang},
  booktitle={Proceedings of the SIGGRAPH Asia 2025 Conference Papers},
  pages={1--11},
  year={2025}
}

@inproceedings{geopard,
  title={Geopard: Geometric pretraining for articulation prediction in 3d shapes},
  author={Goyal, Pradyumn and Petrov, Dmitry and Andrews, Sheldon and Ben-Shabat, Yizhak and Liu, Hsueh-Ti Derek and Kalogerakis, Evangelos},
  booktitle={Proceedings of the IEEE/CVF International Conference on Computer Vision},
  pages={9332--9341},
  year={2025}
}

@inproceedings{freeart3d,
  title={Freeart3d: Training-free articulated object generation using 3d diffusion},
  author={Chen, Chuhao and Liu, Isabella and Wei, Xinyue and Su, Hao and Liu, Minghua},
  booktitle={Proceedings of the SIGGRAPH Asia 2025 Conference Papers},
  pages={1--13},
  year={2025}
}

@article{kinematify,
  title={Kinematify: Open-Vocabulary Synthesis of High-DoF Articulated Objects},
  author={Wang, Jiawei and Wang, Dingyou and Hu, Jiaming and Zhang, Qixuan and Yu, Jingyi and Xu, Lan},
  journal={arXiv preprint arXiv:2511.01294},
  year={2025}
}

@article{articulate-anymesh,
  title={Articulate anymesh: Open-vocabulary 3d articulated objects modeling},
  author={Qiu, Xiaowen and Yang, Jincheng and Wang, Yian and Chen, Zhehuan and Wang, Yufei and Wang, Tsun-Hsuan and Xian, Zhou and Gan, Chuang},
  journal={arXiv preprint arXiv:2502.02590},
  year={2025}
}

@article{atop,
  title={Articulate That Object Part (ATOP): 3D Part Articulation from Text and via Motion Personalization},
  author={Vora, Aditya and Nag, Sauradip and Wang, Kai and Zhang, Hao},
  journal={ACM Transactions on Graphics},
  year={2025},
  publisher={ACM New York, NY}
}

@inproceedings{cage,
  title={Cage: Controllable articulation generation},
  author={Liu, Jiayi and Tam, Hou In Ivan and Mahdavi-Amiri, Ali and Savva, Manolis},
  booktitle={Proceedings of the IEEE/CVF Conference on Computer Vision and Pattern Recognition},
  pages={17880--17889},
  year={2024}
}

@inproceedings{lam,
  title={LAM: Language Articulated Object Modelers},
  author={Gao, Yipeng and Ge, Yunhao and Cai, Peilin and Seita, Daniel and Itti, Laurent},
  booktitle={Proceedings of the IEEE/CVF Conference on Computer Vision and Pattern Recognition},
  pages={16010--16020},
  year={2026}
}

@misc{fgclip2,
      title={FG-CLIP 2: A Bilingual Fine-grained Vision-Language Alignment Model}, 
      author={Chunyu Xie and Bin Wang and Fanjing Kong and Jincheng Li and Dawei Liang and Ji Ao and Dawei Leng and Yuhui Yin},
      year={2026},
      eprint={2510.10921},
      archivePrefix={arXiv},
      primaryClass={cs.CV},
      url={https://arxiv.org/abs/2510.10921}, 
}

@misc{langsplat,
      title={LangSplat: 3D Language Gaussian Splatting}, 
      author={Minghan Qin and Wanhua Li and Jiawei Zhou and Haoqian Wang and Hanspeter Pfister},
      year={2024},
      eprint={2312.16084},
      archivePrefix={arXiv},
      primaryClass={cs.CV},
      url={https://arxiv.org/abs/2312.16084}, 
}

@misc{dff,
      title={Decomposing NeRF for Editing via Feature Field Distillation}, 
      author={Sosuke Kobayashi and Eiichi Matsumoto and Vincent Sitzmann},
      year={2022},
      eprint={2205.15585},
      archivePrefix={arXiv},
      primaryClass={cs.CV},
      url={https://arxiv.org/abs/2205.15585} 
}

@misc{feature3dgs,
      title={Feature 3DGS: Supercharging 3D Gaussian Splatting to Enable Distilled Feature Fields}, 
      author={Shijie Zhou and Haoran Chang and Sicheng Jiang and Zhiwen Fan and Zehao Zhu and Dejia Xu and Pradyumna Chari and Suya You and Zhangyang Wang and Achuta Kadambi},
      year={2024},
      eprint={2312.03203},
      archivePrefix={arXiv},
      primaryClass={cs.CV},
      url={https://arxiv.org/abs/2312.03203}, 
}

@misc{openscene,
      title={OpenScene: 3D Scene Understanding with Open Vocabularies}, 
      author={Songyou Peng and Kyle Genova and Chiyu "Max" Jiang and Andrea Tagliasacchi and Marc Pollefeys and Thomas Funkhouser},
      year={2023},
      eprint={2211.15654},
      archivePrefix={arXiv},
      primaryClass={cs.CV},
      url={https://arxiv.org/abs/2211.15654}, 
}

@inproceedings{clipfields, series={RSS2023},
   title={CLIP-Fields: Weakly Supervised Semantic Fields for Robotic Memory},
   url={http://dx.doi.org/10.15607/RSS.2023.XIX.074},
   DOI={10.15607/rss.2023.xix.074},
   booktitle={Robotics: Science and Systems XIX},
   publisher={Robotics: Science and Systems Foundation},
   author={Shafiullah, Nur Muhammad Mahi and Paxton, Chris and Pinto, Lerrel and Chintala, Soumith and Szlam, Arthur},
   year={2023},
   month=July, collection={RSS2023} }

@misc{garfield,
      title={GARField: Group Anything with Radiance Fields}, 
      author={Chung Min Kim and Mingxuan Wu and Justin Kerr and Ken Goldberg and Matthew Tancik and Angjoo Kanazawa},
      year={2024},
      eprint={2401.09419},
      archivePrefix={arXiv},
      primaryClass={cs.CV},
      url={https://arxiv.org/abs/2401.09419}, 
}

@misc{in2n,
      title={Instruct-NeRF2NeRF: Editing 3D Scenes with Instructions}, 
      author={Ayaan Haque and Matthew Tancik and Alexei A. Efros and Aleksander Holynski and Angjoo Kanazawa},
      year={2023},
      eprint={2303.12789},
      archivePrefix={arXiv},
      primaryClass={cs.CV},
      url={https://arxiv.org/abs/2303.12789}, 
}

@misc{gaussianeditor,
      title={GaussianEditor: Swift and Controllable 3D Editing with Gaussian Splatting}, 
      author={Yiwen Chen and Zilong Chen and Chi Zhang and Feng Wang and Xiaofeng Yang and Yikai Wang and Zhongang Cai and Lei Yang and Huaping Liu and Guosheng Lin},
      year={2023},
      eprint={2311.14521},
      archivePrefix={arXiv},
      primaryClass={cs.CV},
      url={https://arxiv.org/abs/2311.14521}, 
}

@misc{lerf,
      title={LERF: Language Embedded Radiance Fields}, 
      author={Justin Kerr and Chung Min Kim and Ken Goldberg and Angjoo Kanazawa and Matthew Tancik},
      year={2023},
      eprint={2303.09553},
      archivePrefix={arXiv},
      primaryClass={cs.CV},
      url={https://arxiv.org/abs/2303.09553}, 
}

@misc{clip,
      title={Learning Transferable Visual Models From Natural Language Supervision}, 
      author={Alec Radford and Jong Wook Kim and Chris Hallacy and Aditya Ramesh and Gabriel Goh and Sandhini Agarwal and Girish Sastry and Amanda Askell and Pamela Mishkin and Jack Clark and Gretchen Krueger and Ilya Sutskever},
      year={2021},
      eprint={2103.00020},
      archivePrefix={arXiv},
      primaryClass={cs.CV},
      url={https://arxiv.org/abs/2103.00020}, 
}

@inproceedings{regionclip,
  title={Regionclip: Region-based language-image pretraining},
  author={Zhong, Yiwu and Yang, Jianwei and Zhang, Pengchuan and Li, Chunyuan and Codella, Noel and Li, Liunian Harold and Zhou, Luowei and Dai, Xiyang and Yuan, Lu and Li, Yin and others},
  booktitle={Proceedings of the IEEE/CVF Conference on Computer Vision and Pattern Recognition},
  pages={16793--16803},
  year={2022}
}

@misc{groundedsam,
      title={Grounded SAM: Assembling Open-World Models for Diverse Visual Tasks}, 
      author={Tianhe Ren and Shilong Liu and Ailing Zeng and Jing Lin and Kunchang Li and He Cao and Jiayu Chen and Xinyu Huang and Yukang Chen and Feng Yan and Zhaoyang Zeng and Hao Zhang and Feng Li and Jie Yang and Hongyang Li and Qing Jiang and Lei Zhang},
      year={2024},
      eprint={2401.14159},
      archivePrefix={arXiv},
      primaryClass={cs.CV}
}

@article{gpt4o,
  title   = {{GPT-4o} System Card},
  author  = {{OpenAI}},
  journal = {arXiv preprint arXiv:2410.21276},
  year    = {2024}
}
}

\clearpage

\appendix

\section{Additional Experiments}
\label{sec:supp_experiments}
This section reports the full set of additional experiments. Most of the language experiments use a single procedurally generated command dataset whose statistics are summarized in Table~\ref{tab:supp_dataset_stats}, with objects and compositions split disjointly between training and evaluation. The few experiments that use a different set state so explicitly where they appear: the oracle-parser comparison is run on a separate, larger $100$-object generated command set (Table~\ref{tab:supp_oracle_parser}), and the anchor/field ablation is run on constructed synthetic diagnostic scenes (Table~\ref{tab:supp_anchor_field}).

\begin{table}[htbp]
\centering
\caption{Command-grounding dataset statistics by split.}
\label{tab:supp_dataset_stats}
\setlength{\tabcolsep}{4pt}\renewcommand{\arraystretch}{1.1}
\resizebox{\ifdim\width>\linewidth\linewidth\else\width\fi}{!}{
\begin{tabular}{lccccc}
\toprule
Split & \#Obj & \#Cmd & \#Nodes & \% null & \% rep-part \\
\midrule
Train & 70 & 6576 & 10770 & 3.9 & 64.6 \\
Val & 15 & 1362 & 2220 & 4.1 & 67.2 \\
Test & 15 & 1152 & 1836 & 4.9 & 62.3 \\
Total & 100 & 9090 & 14826 & 4.0 & 64.7 \\
\bottomrule
\end{tabular}
}
\end{table}

\paragraph{Structured directive-to-part assignment.}
Table~\ref{tab:supp_grounding_ablations} isolates the global assignment by comparing it against independent per-directive selection and a variant without the relation term, both on the full command set and on a contrastive subset of sibling-collision and relational-chain commands constructed so that independent selection is forced into an error. The joint assignment improves accuracy on every subset, with the largest gains on exactly the collision and relational cases the formulation is designed for. Table~\ref{tab:supp_oracle_parser} substitutes oracle directive graphs for the rule-based parser: grounding on the two graphs differs by less than the run-to-run variation, so on this command distribution the parser is not the dominant error source. The three grounding tables are not directly comparable cell-for-cell: Table~\ref{tab:supp_grounding_ablations} scores every directive of the $15$-object \texttt{data\_cmd} test split, Table~\ref{tab:supp_oracle_parser} scores a separate $100$-object procedurally generated command set, and the entity-to-part row of Table~\ref{tab:supp_command_breakdown} counts every directive grounded to a real ground-truth part, including the reference-only anchor entities the resolver must also place, and excluding only null and ambiguous directives; its denominator therefore exceeds that of the verb row, which counts only actionable verb-bearing directives, by exactly the reference-only anchors.

\begin{table}[htbp]
\centering
\caption{Directive-to-part assignment accuracy: global graph assignment vs.\ independent \texttt{argmax} and a NoRelation ablation, by subset, on the synthetic command dataset.}
\label{tab:supp_grounding_ablations}
\setlength{\tabcolsep}{4pt}\renewcommand{\arraystretch}{1.1}
\resizebox{\ifdim\width>\linewidth\linewidth\else\width\fi}{!}{
\begin{tabular}{lcccc}
\toprule
Method & All $\uparrow$ & Multi-directive $\uparrow$ & Relational $\uparrow$ & Hard $\uparrow$ \\
\midrule
Global & \textbf{0.842} & \textbf{0.899} & \textbf{0.827} & \textbf{0.790} \\
Independent & 0.783 & 0.812 & 0.708 & 0.543 \\
NoRelation & 0.789 & 0.812 & 0.701 & 0.530 \\
\bottomrule
\end{tabular}
}
\end{table}

\begin{table}[htbp]
\centering
\caption{Grounding accuracy with the rule-based parser vs.\ oracle directive graphs, isolating parser error.}
\label{tab:supp_oracle_parser}
\setlength{\tabcolsep}{4pt}\renewcommand{\arraystretch}{1.1}
\resizebox{\ifdim\width>\linewidth\linewidth\else\width\fi}{!}{
\begin{tabular}{lcc}
\toprule
Parser & All $\uparrow$ & Multi-directive $\uparrow$ \\
\midrule
Predicted (rule parser) & \textbf{0.870} & \textbf{0.909} \\
Oracle & 0.863 & 0.901 \\
\bottomrule
\end{tabular}
}
\end{table}

\paragraph{Per-stage accuracy and dispersion.}
Table~\ref{tab:supp_command_breakdown} decomposes command grounding into its stages: parse count agreement (the parse has the same actionable-directive and relation-edge counts as the oracle graph), entity-to-part accuracy, verb accuracy, magnitude-\emph{type} accuracy over four classes (endpoint, absolute-metric, relative-metric, and fraction; of the five request types of the typed target conversion in the main paper the absolute- and relative-fraction cases are merged into one fraction class, and only the type is scored, not the numeric amount), whole-command success where every directive in a command must be correct, and correct abstention on ambiguous commands. Because the object-disjoint test split contains few objects, Table~\ref{tab:supp_command_dispersion} additionally reports the per-object macro mean, standard deviation, and $95\%$ confidence interval for each stage, computed over the objects that contain each metric's instances.

\begin{table}[htbp]
\centering
\caption{Per-stage accuracy breakdown of the command pipeline on the command dataset (test split, trained resolver $\theta$). \emph{Parse (count agreement)}: fraction of commands whose rule parse has the same actionable-directive and relation-edge counts as the oracle graph (a necessary condition, not a full labeled-graph match, so it upper-bounds true graph recovery). \emph{Entity-to-part}: fraction of every directive grounded to a real ground-truth part (all real-part directives, \emph{including} the reference-only anchor entities the resolver must also place; only null and ambiguous directives are excluded) that the global resolver grounds correctly ($Y_i^\star=Y_{i,gt}$), on the oracle parse. \emph{Verb}: fraction of actionable (verb-bearing) directives whose parser-recovered verb (canonicalised to open/close) matches ground truth; its $n$ is smaller than the Entity-to-part $n$ by exactly the reference-only anchors, which carry a part but no verb. \emph{Magnitude}: fraction of typed-magnitude probes whose recovered magnitude type is correct over four classes (endpoint, absolute-metric, relative-metric, and fraction; the absolute- and relative-fraction requests are merged into one fraction class, and only the type is scored, not the numeric amount). \emph{Complete-command}: fraction of non-ambiguous commands the end-to-end pipeline (predicted parse $+$ grounding) gets entirely right. \emph{Correct abstention}: fraction of ambiguous bare-noun directives ($Y_{gt}=-2$) on which the resolver correctly abstains. $n$ is the number of scored instances.}
\label{tab:supp_command_breakdown}
\setlength{\tabcolsep}{4pt}\renewcommand{\arraystretch}{1.1}
\resizebox{\ifdim\width>\linewidth\linewidth\else\width\fi}{!}{
\begin{tabular}{lcc}
\toprule
Stage & Accuracy $\uparrow$ & $n$ \\
\midrule
Directive-graph parse & 0.844 & 1152 \\
Entity-to-part & 0.913 & 1686 \\
Verb & 0.889 & 1566 \\
Magnitude & 1.000 & 152 \\
Complete-command & 0.673 & 1092 \\
Correct abstention (ambiguous) & 0.200 & 60 \\
\bottomrule
\end{tabular}
}
\end{table}

\begin{table}[htbp]
\centering
\caption{Per-object dispersion of the command-pipeline metrics across the test objects; because the object-disjoint test split is small we quantify uncertainty. We report, for each metric, the macro mean $\pm$ standard deviation over the per-object accuracies and a 95\% Student's-$t$ confidence interval, where $n$ is the number of test objects that contain that metric's instances and varies by row; the intervals are truncated to $[0,1]$ for display. Metrics are computed exactly as in Table~\ref{tab:supp_command_breakdown}.}
\label{tab:supp_command_dispersion}
\setlength{\tabcolsep}{4pt}\renewcommand{\arraystretch}{1.1}
\resizebox{\ifdim\width>\linewidth\linewidth\else\width\fi}{!}{
\begin{tabular}{lccc}
\toprule
Metric & Mean $\uparrow$ & Std & 95\% CI \\
\midrule
Directive-graph parse & 0.779 & 0.098 & [0.725, 0.834] \\
Entity-to-part & 0.958 & 0.081 & [0.913, 1.000] \\
Verb & 0.805 & 0.112 & [0.743, 0.867] \\
Magnitude & 1.000 & 0.000 & [1.000, 1.000] \\
Complete-command & 0.554 & 0.184 & [0.452, 0.656] \\
Correct abstention (ambiguous) & 0.222 & 0.441 & [0.000, 0.561] \\
\bottomrule
\end{tabular}
}
\end{table}

\begin{figure}[htbp]
  \centering
  \includegraphics[width=0.8\linewidth]{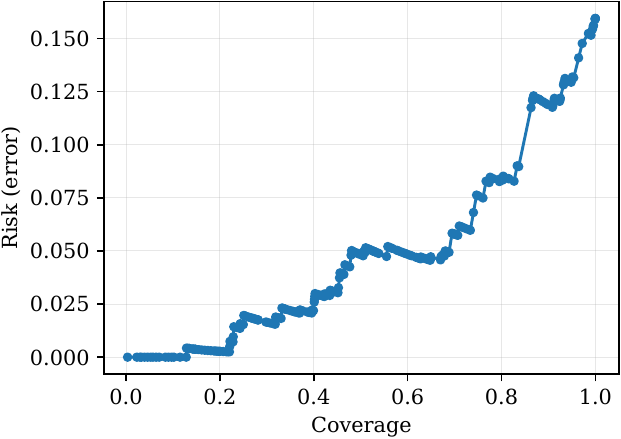}
  \caption{Risk-coverage curve for structured-margin abstention. Abstaining on the lowest-margin directives lowers coverage and reduces selective risk.}
  \label{fig:supp_riskcov}
\end{figure}

\paragraph{Abstention and ambiguity.}
Table~\ref{tab:supp_ambiguity} reports accuracy with and without the abstention mechanism on an ambiguity-weighted subset in which should-abstain directives are $30\%$ of the total. Holding the global resolver fixed and toggling only the null and margin gate, abstention raises accuracy from $64.2\%$ to $68.2\%$ ($\Delta=+4.0\%$). Figure~\ref{fig:supp_riskcov} plots the risk-coverage curve for structured-margin selective prediction on the full test distribution, whose area under the curve is 0.048 (lower is better). The global assignment solves in sub-second time, a few milliseconds across all tested part and directive counts, so solve time is not a runtime bottleneck.

\begin{table}[htbp]
\centering
\caption{Abstention accuracy on an ambiguity stress subset: every should-abstain directive (ambiguous or absent-referent) plus a random, seed-fixed sample of resolvable directives, so should-abstain is 30\% of the 500 scored directives (the full command distribution is only a few percent should-abstain, which bounds the achievable gain). The global resolver is held fixed and only the abstention gate is toggled: \emph{Acc (no abstain)} forces the global assignment to always execute, \emph{Acc (with abstain)} adds the null/margin gate, and $\Delta$ is the abstention gain.}
\label{tab:supp_ambiguity}
\setlength{\tabcolsep}{4pt}\renewcommand{\arraystretch}{1.1}
\resizebox{\ifdim\width>\linewidth\linewidth\else\width\fi}{!}{
\begin{tabular}{lc}
\toprule
Metric & Value \\
\midrule
Acc (no abstain) & 0.642 \\
Acc (with abstain) & 0.682 \\
$\Delta$ (abstention gain) & 0.040 \\
Should-abstain fraction & 0.300 \\
\bottomrule
\end{tabular}
}
\end{table}

\begin{figure}
  \centering
  \includegraphics[width=0.8\linewidth]{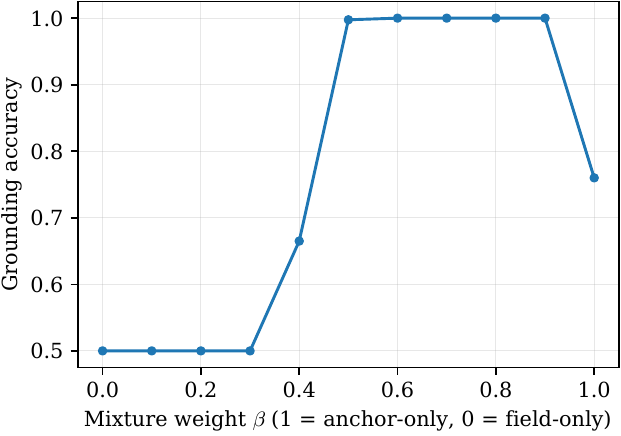}
  \caption{Grounding accuracy as the semantic energy is mixed between field-only ($\beta{=}0$) and anchor-only ($\beta{=}1$); a broad combined region improves over either endpoint.}
  \label{fig:supp_anchor_field}
\end{figure}

\paragraph{Semantic anchor versus dense field.}
The semantic energy combines a discrete anchor term with a dense distilled field. Table~\ref{tab:supp_anchor_field} compares anchor-only, field-only, and the combined energy on two synthetic diagnostic probes of $200$ constructed scenes each, one where resolution is decided by the anchor and one where it is decided by the field, and neither term alone handles both while the combined energy does; Figure~\ref{fig:supp_anchor_field} plots grounding accuracy as the mixing weight is swept, where a broad combined region improves over either endpoint.

\begin{table}[htbp]
\centering
\caption{Anchor / field / anchor+field ablation of the semantic score. Anchor-only ($\beta{=}1$) cannot separate same-noun siblings; field-only ($\beta{=}0$) is misled by a distractor part whose dense field spuriously matches the query; a broad range of mixtures around the default ($\beta{=}0.6$) handles both, so the distilled dense field earns its place. These are diagnostic probes over constructed anchor- and field-decisive cases, not the full metric set.}
\label{tab:supp_anchor_field}
\setlength{\tabcolsep}{4pt}\renewcommand{\arraystretch}{1.1}
\resizebox{\ifdim\width>\linewidth\linewidth\else\width\fi}{!}{
\begin{tabular}{lccc}
\toprule
Semantic score & Anchor-decisive $\uparrow$ & Field-decisive $\uparrow$ & All $\uparrow$ \\
\midrule
Anchor-only ($\beta{=}1$) & \textbf{1.000} & 0.520 & 0.760 \\
Field-only ($\beta{=}0$) & 0.000 & \textbf{1.000} & 0.500 \\
Both ($\beta{=}0.6$) & \textbf{1.000} & \textbf{1.000} & \textbf{1.000} \\
\bottomrule
\end{tabular}
}
\end{table}

\begin{figure*}[tp]
  \centering
  \includegraphics[width=0.9\textwidth]{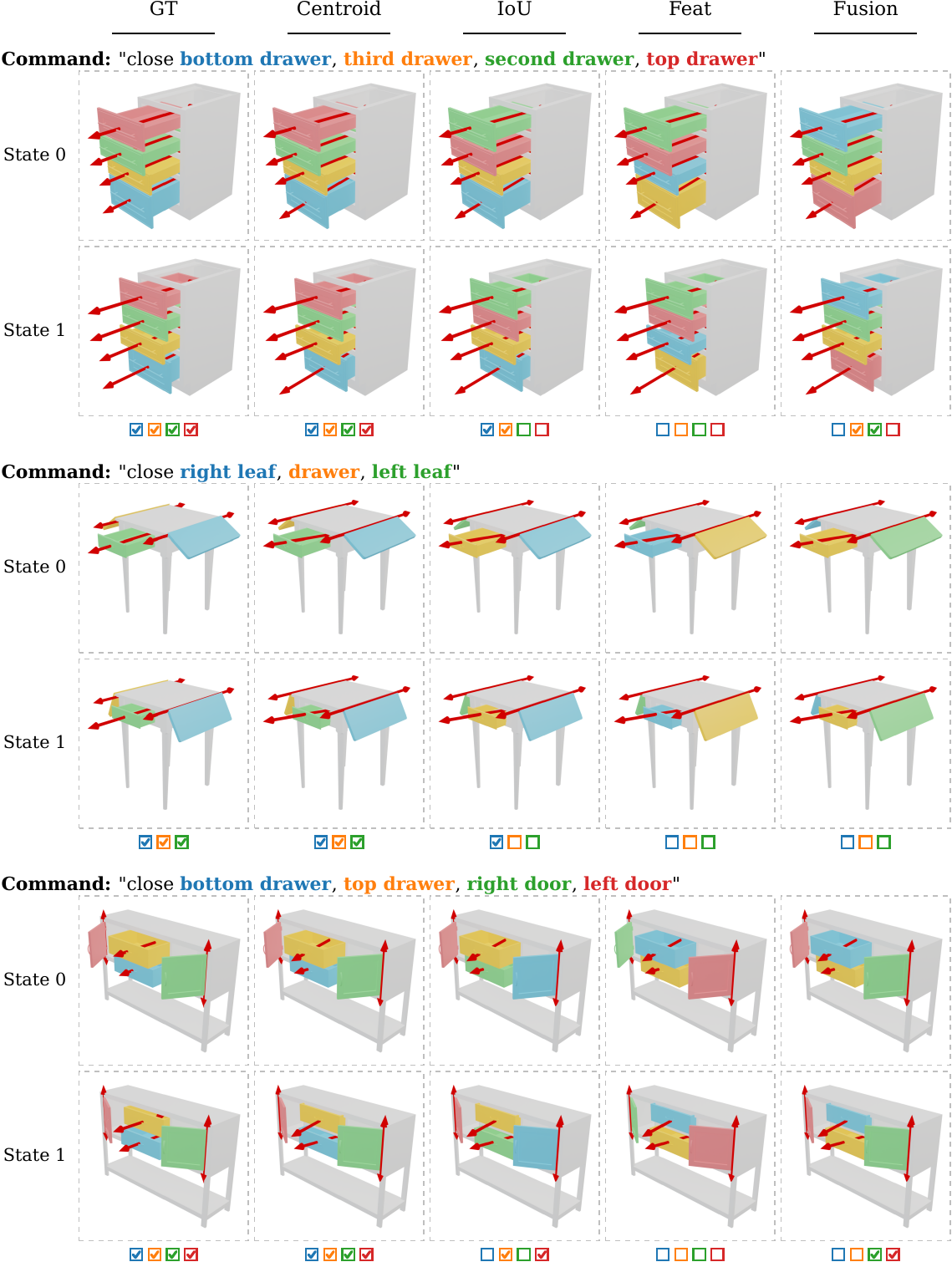}
  \caption{Part-binding cost ablation, qualitatively. The same geometry is recolored by the centroid, mask-IoU, feature-similarity, and learned namers, with per-part correctness marks.}
  \label{fig:supp_qual_binding}
\end{figure*}

\begin{figure}[htbp]
  \centering
  \includegraphics[width=0.8\linewidth]{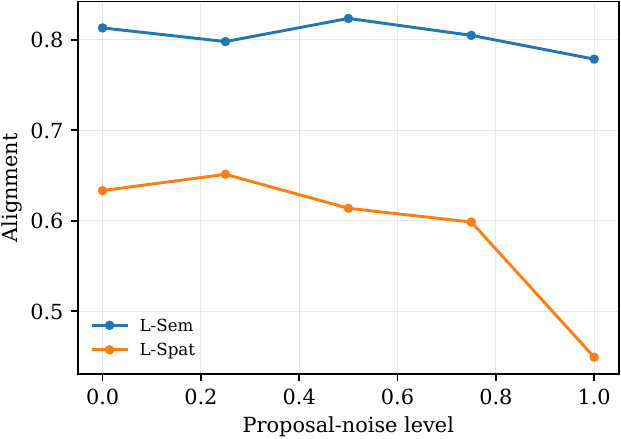}
  \caption{Naming metrics as the vision-language proposals are perturbed with increasing noise.}
  \label{fig:supp_proposal_noise}
\end{figure}

\begin{figure*}[tp]
  \centering
  \includegraphics[width=0.8\textwidth]{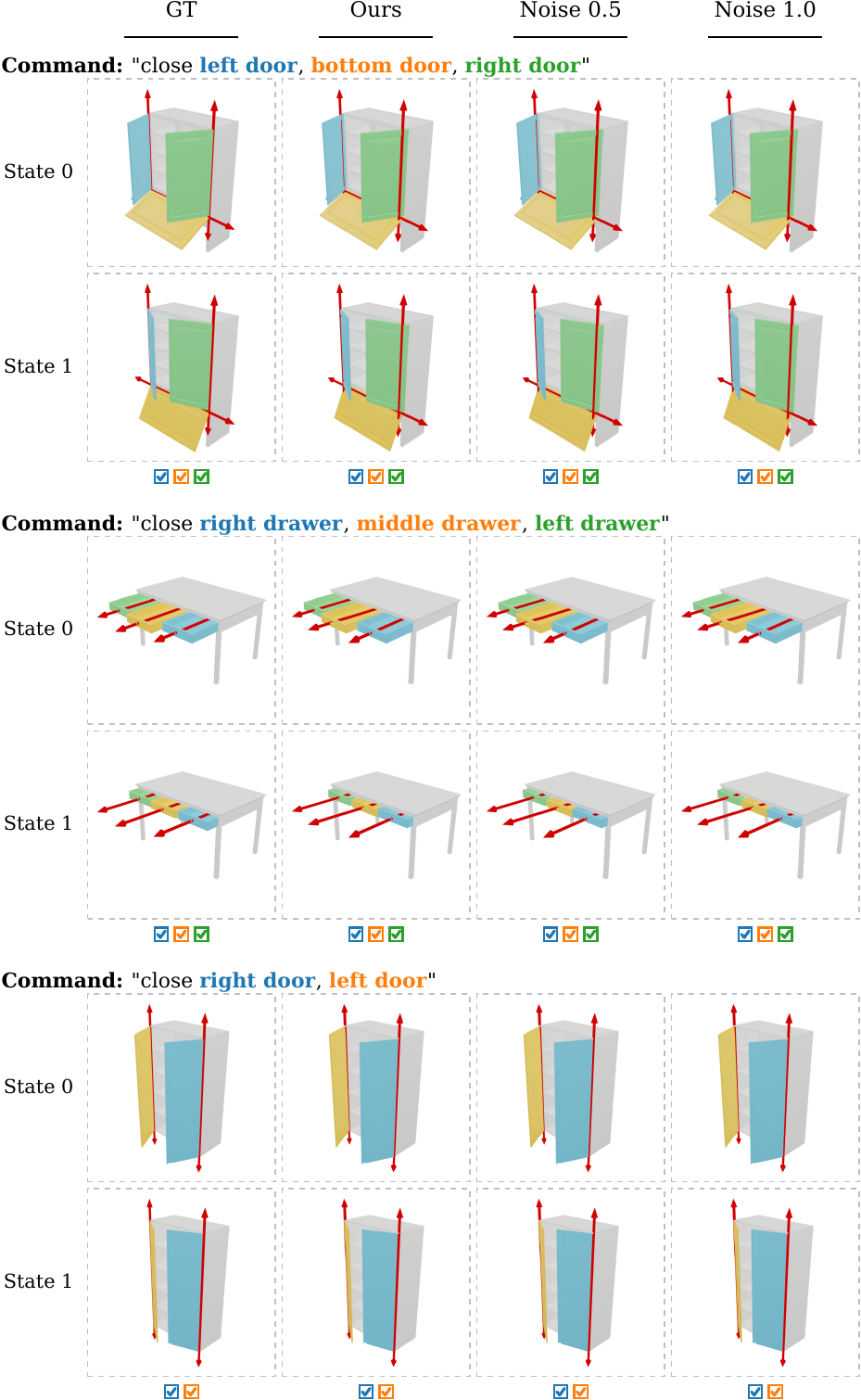}
  \caption{Part naming under proposal noise, qualitatively, at increasing perturbation levels.}
  \label{fig:supp_qual_proposal}
\end{figure*}

\begin{figure*}[tp]
  \centering
  \includegraphics[width=0.9\textwidth]{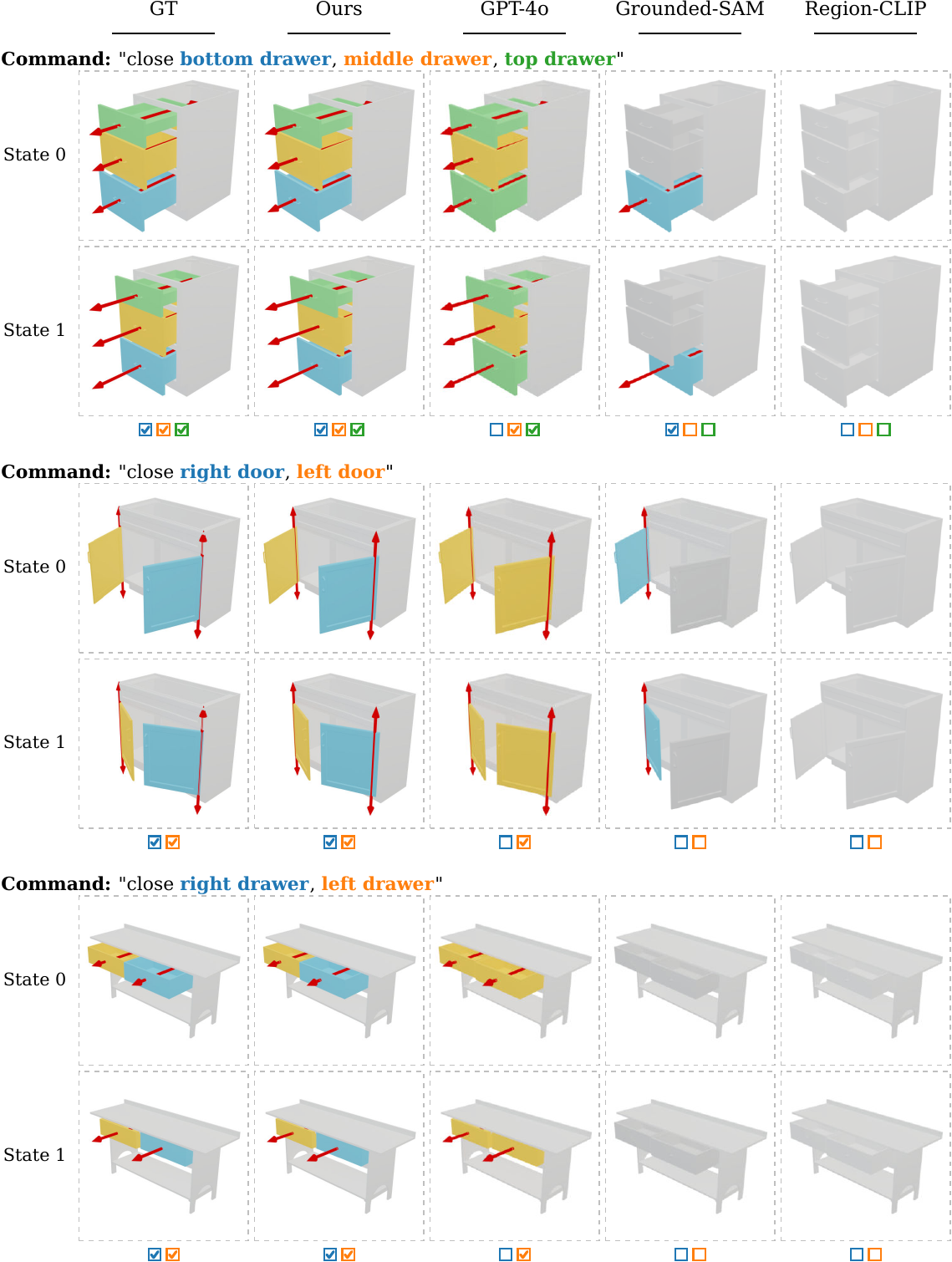}
  \caption{Language-grounding baselines, qualitatively. The same geometry is recolored by each baseline's predicted naming, with per-part correctness marks.}
  \label{fig:supp_qual_lang}
\end{figure*}

\paragraph{Part naming and binding.}
Table~\ref{tab:supp_binding_variants} ablates the part-binding cost, comparing the projected-centroid assignment used by our method against mask overlap, feature similarity, and a learned matcher, with the corresponding qualitative recolorings in Figure~\ref{fig:supp_qual_binding}. Figure~\ref{fig:supp_proposal_noise} stresses the naming pipeline by perturbing the vision-language proposals and re-deriving names over a fixed object subsample, plotting the language metrics against noise; the overall trend is downward, though the first perturbation levels can rise before falling rather than decreasing monotonically, and the zero-noise point reflects that subsample rather than the full-test binding numbers of Table~\ref{tab:supp_binding_variants}. Qualitative examples are in Figure~\ref{fig:supp_qual_proposal}. Table~\ref{tab:supp_lang_baselines} compares the alignment metrics against region-CLIP~\cite{regionclip} and Grounded-SAM~\cite{groundedsam}, with qualitative comparisons in Figure~\ref{fig:supp_qual_lang}. We also probe a directly prompted vision-language model: GPT-4o~\cite{gpt4o}, shown the rendered object with a numbered legend of its parts and asked to name the part each command acts on, grounds correctly on only $27.3\%$ of the $282$ answered commands. These $282$ are the \emph{basic} single-referent commands, one \emph{close the $\langle$part$\rangle$} command per movable part, not the relational or compound directives grounded elsewhere, so even on the easiest slice of the task a prompted model trails the structured resolver by a wide margin.

\begin{table}[htbp]
\centering
\caption{Part-binding cost ablation: language metrics under centroid, mask-IoU, feature-similarity, and learned (fusion-head) matching. The underlying geometry and joint trajectories are fixed across namers, so pure reconstruction metrics are omitted; \textsc{L-M} is retained because naming changes the correspondence used by the language-motion metric.}
\label{tab:supp_binding_variants}
\setlength{\tabcolsep}{4pt}\renewcommand{\arraystretch}{1.1}
\resizebox{\ifdim\width>\linewidth\linewidth\else\width\fi}{!}{
\begin{tabular}{lccc}
\toprule
Namer & L-Sem $\uparrow$ & L-Spat $\uparrow$ & L-M $\downarrow$ \\
\midrule
Centroid & 0.775 & \textbf{0.607} & \textbf{70.42} \\
IoU & \textbf{0.797} & 0.547 & 86.32 \\
Feat & 0.784 & 0.095 & 326.32 \\
Fusion & 0.725 & 0.413 & 209.00 \\
\bottomrule
\end{tabular}
}

\end{table}

\begin{table}[htbp]
\centering
\caption{Language-grounding baselines vs.\ ArtLang on the semantic and spatial alignment axes.}
\label{tab:supp_lang_baselines}
\setlength{\tabcolsep}{4pt}\renewcommand{\arraystretch}{1.1}
\resizebox{\ifdim\width>\linewidth\linewidth\else\width\fi}{!}{
\begin{tabular}{lcc}
\toprule
Method & L-Sem $\uparrow$ & L-Spat $\uparrow$ \\
\midrule
Region-CLIP & 0.335 & 0.084 \\
Grounded-SAM & 0.538 & 0.180 \\
ArtLang (Ours) & \textbf{0.775} & \textbf{0.607} \\
\bottomrule
\end{tabular}
}
\end{table}

\paragraph{Intent preservation under paraphrase.}
The metrics above charge the templated parser for recovering the exact surface noun, which understates how the interface behaves in use, where a referent is grounded to a part by embedding rather than by string. Table~\ref{tab:supp_nl_normalize} instead measures whether a language-model paraphrase of each command, introducing reordering, anaphora, and out-of-template vocabulary, still yields the intended \emph{action} and \emph{referent}: the referent is counted correct when it preserves every spatial qualifier and names the same part by string or embedding synonymy, as the semantic anchor grounds it. The parser preserves the intended action-and-referent on the large majority of paraphrases, so the templated interface degrades gracefully to free-form phrasing without any change to the grounding stage. This probe uses a local command sample rather than the pod command split, so its absolute rates are not directly comparable cell-for-cell with the grounding tables.

\begin{table}[htbp]
\centering
\caption{Intent preservation on free-form paraphrases. Language-model paraphrases of the templated commands (with reordering, anaphora, and out-of-template vocabulary) are read by the typed parser and scored against the intent of the original command by whether the resulting directive would drive the same part the same way: \emph{Action} is the verb class, \emph{Referent} requires every spatial qualifier to be preserved and the head noun to match the intended part by string or embedding synonymy (as the semantic anchor grounds it), \emph{Magnitude} is the request type, and \emph{Intent} is Action and Referent together. $n$ is the number of paraphrases scored.}
\label{tab:supp_nl_normalize}
\setlength{\tabcolsep}{4pt}\renewcommand{\arraystretch}{1.1}
\resizebox{\ifdim\width>\linewidth\linewidth\else\width\fi}{!}{
\begin{tabular}{lc}
\toprule
Field & Accuracy $\uparrow$ \\
\midrule
Action (verb class) & 0.864 \\
Referent (part + qualifier) & 0.947 \\
Magnitude type & 0.959 \\
Intent (action $+$ referent) & 0.835 \\
\bottomrule
\end{tabular}
}
\end{table}

\paragraph{Online runtime and endpoint equivariance.}
Table~\ref{tab:supp_runtime} reports the \emph{online} per-stage query latency of the language-to-actuation pipeline; the offline per-asset reconstruction and feature-field distillation are excluded and dominate the total cost. Table~\ref{tab:supp_endpoint} verifies that the endpoint classifier is permutation-equivariant: swapping the observed state ordering flips the predicted index on every scored part (a flip rate of $1.0$; an order-driven predictor that always picked a fixed slot would give $0$), confirming that which state is open is decided from appearance rather than from observation order, as in the endpoint classifier of the main paper.

\begin{table}[htbp]
\centering
\caption{Per-stage runtime of the language-to-actuation pipeline, mean over commands (parser, structured grounding, actuation; CPU). Reconstruction and per-asset feature distillation are one-time offline costs incurred before any command is issued and are excluded from this online timing.}
\label{tab:supp_runtime}
\setlength{\tabcolsep}{4pt}\renewcommand{\arraystretch}{1.1}
\resizebox{\ifdim\width>\linewidth\linewidth\else\width\fi}{!}{
\begin{tabular}{lc}
\toprule
Stage & Time (ms) \\
\midrule
Parse & 0.044 \\
Ground & 3.979 \\
Actuate & 0.009 \\
\bottomrule
\end{tabular}
}
\end{table}

\begin{table}[htbp]
\centering
\caption{Endpoint (open/closed) classifier permutation-equivariance. \emph{Protocol:} the identity of the physical open state is held fixed while its slot changes under the swap, so its correct index flips; the two observed reconstructed states are exchanged and the reported quantity is the \emph{prediction flip rate}, the fraction of parts whose returned open-state index changes, not an accuracy. A high forward accuracy together with a flip rate of $1.0$ is the intended result: the classifier is permutation-equivariant, returning the index of the same appearance-based choice under either input ordering (a genuinely order-driven predictor that always picks a fixed slot would instead have a flip rate of $0$). This is an architectural consistency check, not evidence of learned open/closed semantics. The margin is invariant to the swap.}
\label{tab:supp_endpoint}
\setlength{\tabcolsep}{4pt}\renewcommand{\arraystretch}{1.1}
\resizebox{\ifdim\width>\linewidth\linewidth\else\width\fi}{!}{
\begin{tabular}{lc}
\toprule
Metric & Value \\
\midrule
Original-order accuracy $\uparrow$ & 1.000 \\
Prediction flip rate (states swapped) & 1.000 \\
Mean margin (swap-invariant) & 0.462 \\
\bottomrule
\end{tabular}
}
\end{table}

\subsection{Failure Mode Analysis}
\label{sec:supp_failure}
The most common failure mode of the full pipeline is not in the grounding or actuation stages but upstream, in the open-vocabulary proposals themselves. Our part binding commits a part to the phrase supplied by the vision-language model, and when that model mislabels a part the wrong noun propagates into the semantic anchor. A representative case is a cabinet whose left panel is a hinged door but is proposed by the model as a \emph{drawer}: the geometry, segmentation, and recovered revolute joint are all correct, yet the part is named \emph{drawer}, so a command that asks to \emph{open the door} may bind to the wrong part or abstain. The partial-matching stage and the private dummy nodes limit the damage, because a part can remain unnamed rather than accept an incompatible proposal. When the model returns a confident, geometrically plausible wrong noun, the discrete bound name remains incorrect and no downstream stage can repair the stored name. Command grounding itself can still be rescued in some cases, because the semantic energy also carries a dense distilled field alongside the discrete anchor, together with the spatial and action terms, and the anchor/field ablation in the supplementary material shows the combined score can succeed where either the anchor or the field alone fails; but these terms cannot reliably correct the naming error, and they do not always recover the intended part.

This dependence is a strength as much as a limitation. The semantic quality of \artlang{} is bounded by the quality of the vision-language proposals, and that component is frozen and external: as stronger open-vocabulary models replace the current one, the naming errors shrink without any change to our grounding, actuation, or reconstruction. In this sense \artlang{} inherits improvements in vision-language models for free, and the failures reported here are expected to diminish as those models improve.

\subsection{Additional Qualitative Results}
\label{sec:supp_qual}
The figures in the preceding sections show the articulated end state to save space. Here we provide two-state comparisons for a representative subset of objects, with both the rest state and the articulated state side by side, and the per-part actuation sweep.

\clearpage

\begin{figure*}[t]
  \centering
  \includegraphics[width=\textwidth]{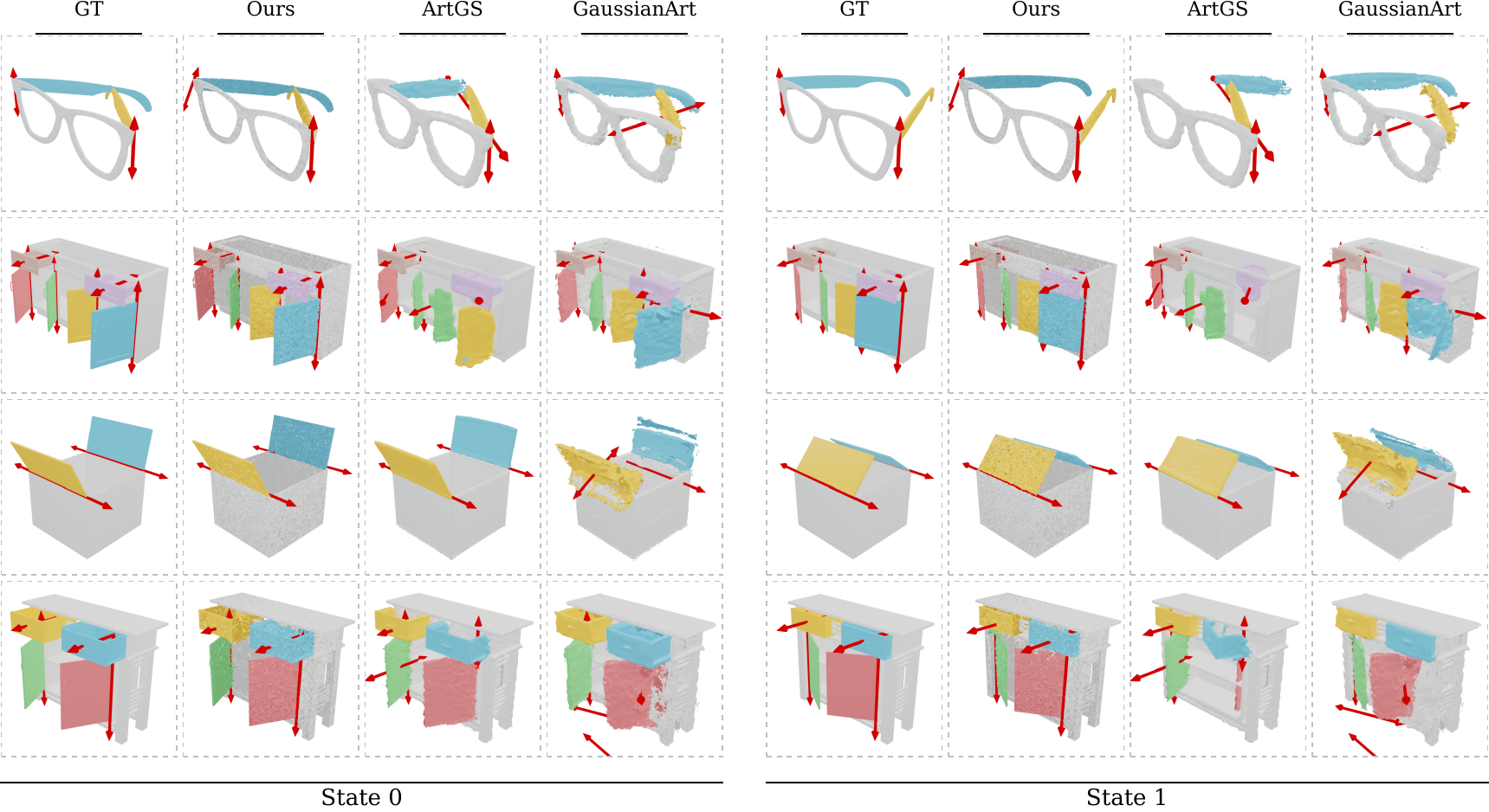}
  \caption{Reconstruction comparison on Articulate-100, both states, for a representative subset of objects. Each row is an object; the two state blocks show the rest and articulated states, with columns for the ground truth, our reconstruction, and the Gaussian baselines, including the per-part segmentation and the joint motion overlay.}
  \label{fig:supp_recon}
\end{figure*}

\begin{figure*}[t]
  \centering
  \includegraphics[width=0.9\textwidth]{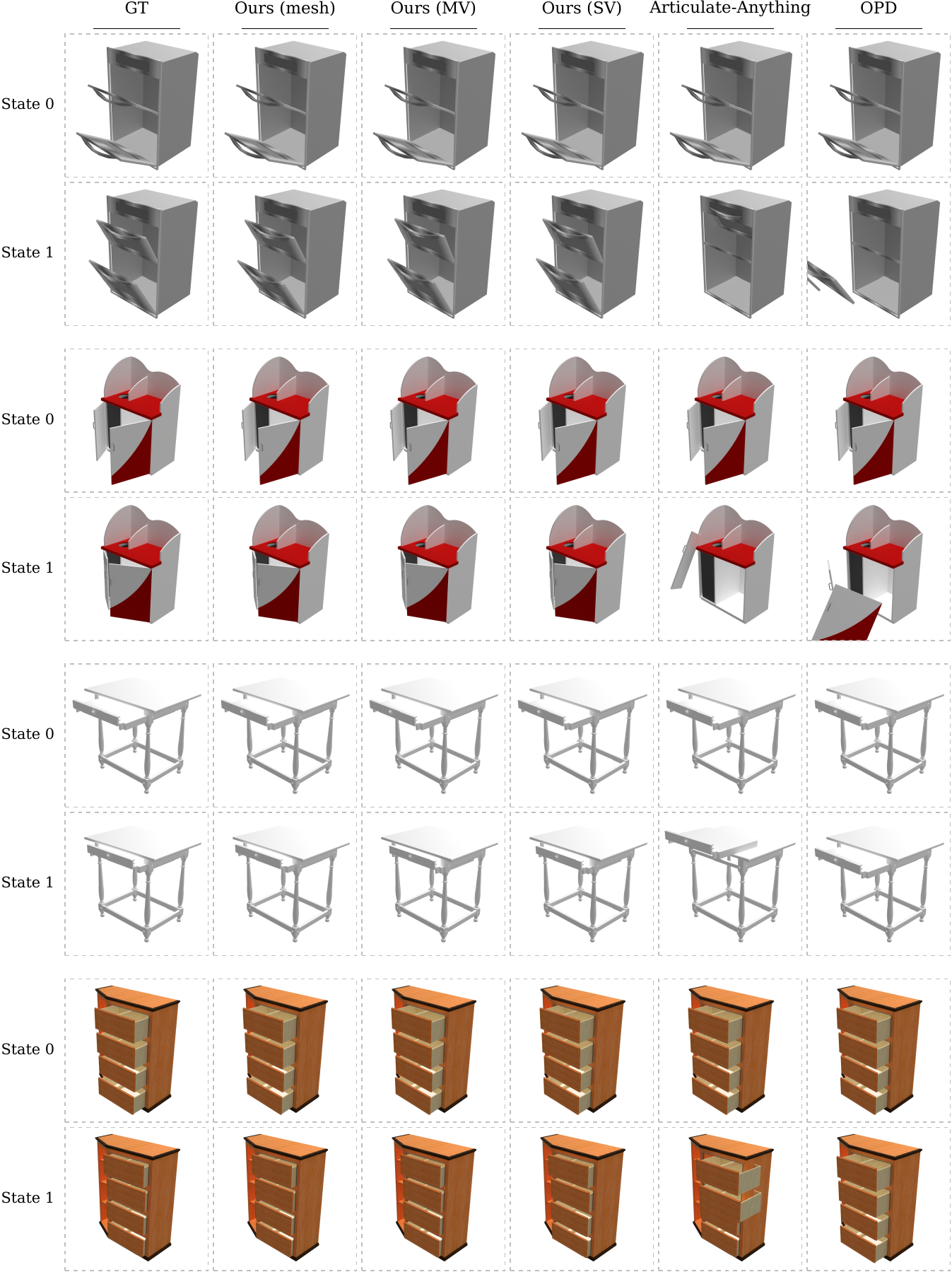}
  \caption{Mesh-input comparison on Articulate-100, both states, for a representative subset of objects. Columns show the ground truth and our mesh-input, multi-view, and single-view variants next to Articulate-Anything~\cite{articulate-anything} and OPD~\cite{opd}, rendered with texture in both the rest and articulated states.}
  \label{fig:supp_mesh}
\end{figure*}

\begin{figure*}[t]
  \centering
  \includegraphics[width=0.65\textwidth]{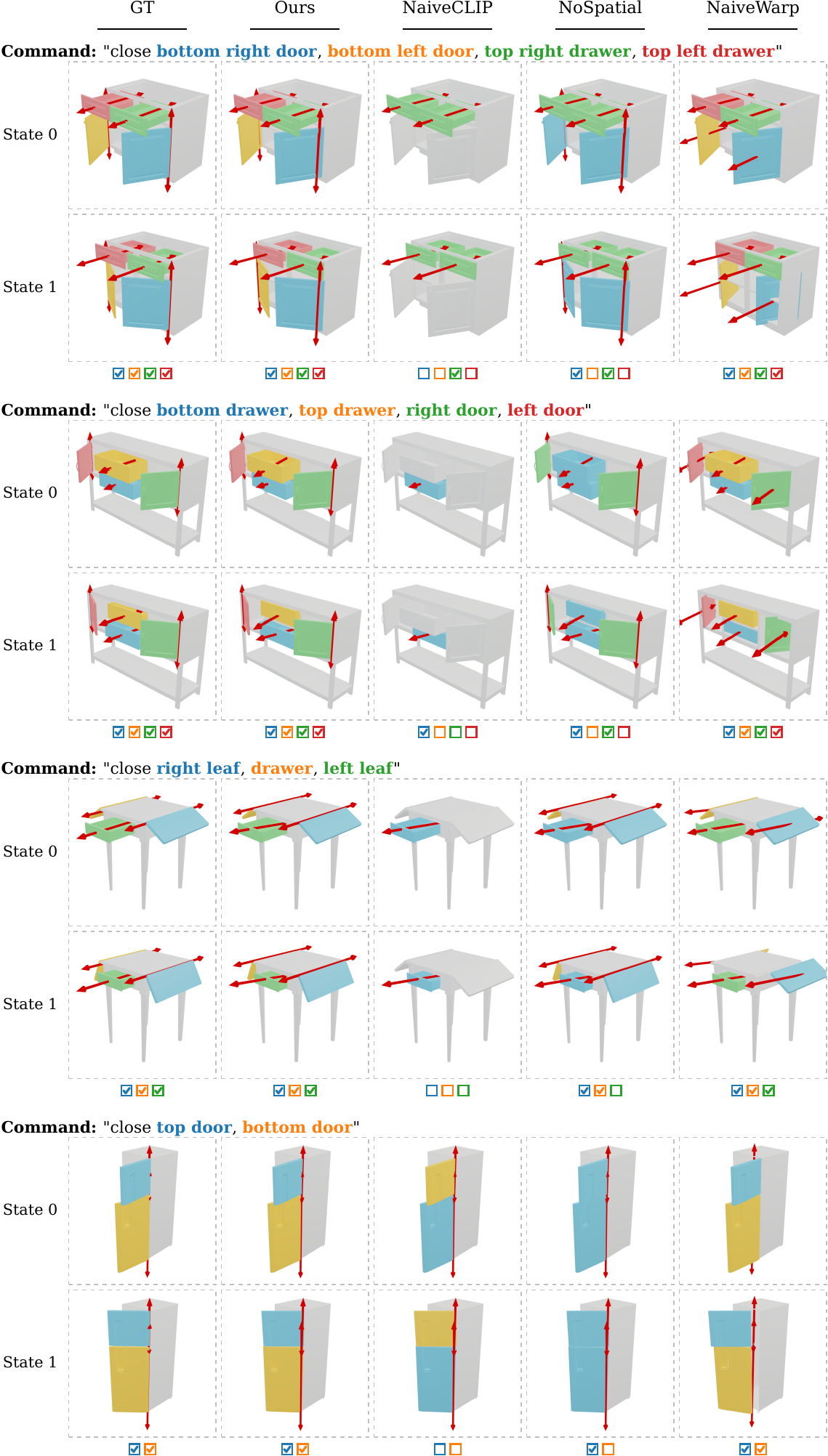}
  \caption{Ablation comparison, both states, for a representative subset of objects. Columns show the ground truth, our model, and the NaiveCLIP, NoSpatial, and NaiveWarp ablations on the same given mesh, in the rest and articulated states.}
  \label{fig:supp_ablation}
\end{figure*}

\begin{figure*}[t]
  \centering
  \includegraphics[width=\textwidth]{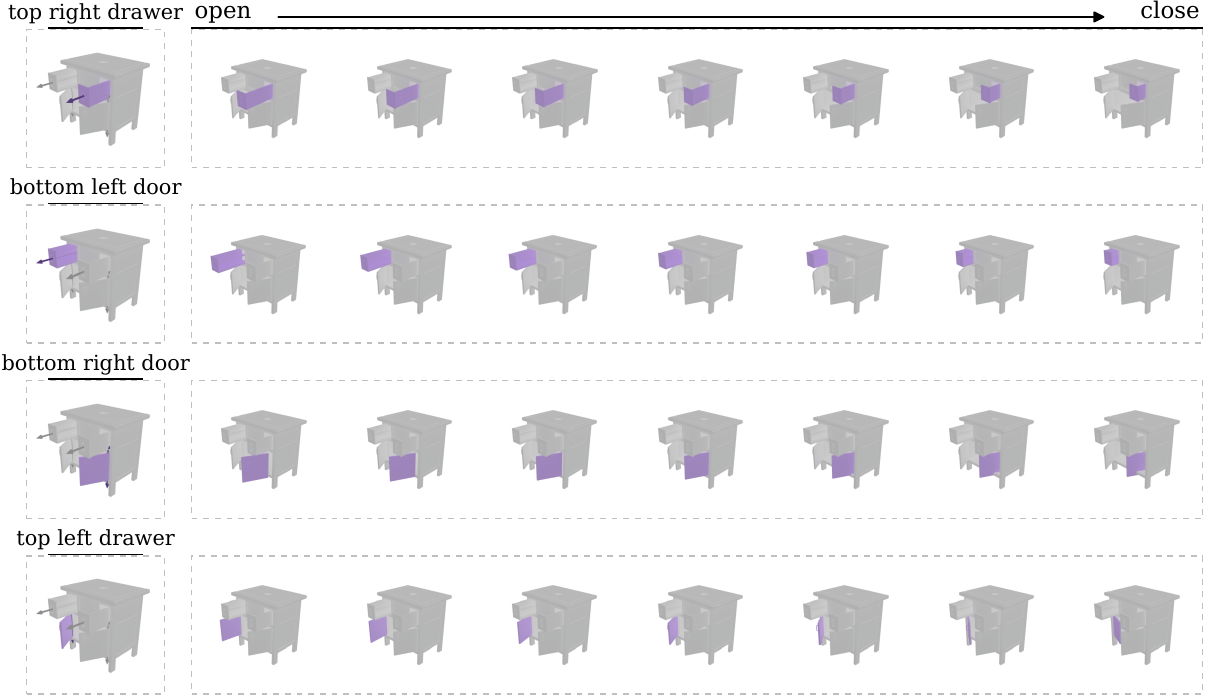}
  \caption{Full continuous actuation sweep. Every movable part of the object is driven along its learned trajectory in turn, one row per part, while the remaining parts are held at rest, sweeping the signed linear factor $s$ of Eq.~\eqref{eq:linear_fraction} from $-1$ to $1$: $s\in[0,1]$ reproduces the observed state $0\!\to\!1$ configurations (for this displayed example state $0$ happens to be the more-open pose, so left-to-right reads as open-to-close; the endpoint classifier does not assume or use this ordering and decides which state is open from appearance), and $s<0$ extrapolates the learned joint past the observed rest state. The one-row version in Figure~\ref{fig:text_move} shows a single part of this figure.}
  \label{fig:supp_text_move}
\end{figure*}

\clearpage


\end{document}